\documentclass{article}
\usepackage{subfig}
\usepackage{graphicx}
\usepackage{amsfonts}
\usepackage{hyperref}
\usepackage{amsmath}
\usepackage{cite}
\usepackage[margin=1in]{geometry}
\title{A systematic dataset generation technique applied to data-driven automotive 
aerodynamics}
\author{Mark Benjamin and Gianluca Iaccarino}

\date{\today}

\begin{document}

\maketitle

\begin{abstract}

    A novel strategy for generating datasets is developed within the context
    of drag prediction for automotive geometries using neural networks. A 
    primary challenge in this space is constructing a training databse of 
    sufficient size and diversity. Our
    method relies on a small number of starting data points, and provides a 
    recipe to interpolate systematically between them, generating an arbitrary
    number of samples at the desired quality. We test this strategy using a
    realistic automotive geometry, and demonstrate that convolutional neural
    networks perform exceedingly well at predicting drag coefficients and 
    surface pressures. Promising results are obtained in testing extrapolation
    performance. Our method can be applied to other problems of 
    aerodynamic shape optimization. 

\end{abstract}

\section{Introduction}
\subsection{Motivation}

The computation of aerodynamic drag is an expensive component of automotive 
design, requiring one computational fluid dynamics simulation per design point. 
It is also an essential 
component of the design process; since drag force is quadratic in vehicle 
velocity, it becomes the dominant mode of energy loss at high speed, affecting 
vehicle top speed, fuel efficiency and stability. The complexity of modern 
vehicle designs leads to a high-dimensional parameter space that can be 
prohibitively expensive to sample. In addition, the evaluation of the quantity 
of interest — the drag coefficient — for each single design point is costly, 
requiring the solution of the Navier-Stokes equations at high Reynolds
numbers, on a mesh sufficiently fine so as to resolve the details of the 
automotive design. Surrogate models are therefore 
of interest; in particular, given the nature of the governing equations and the 
complexity of the geometries of interest, one seeks data-driven models for drag 
prediction. The drag coefficient depends on flow and geometry parameters; 
in our study the flow parameters are fixed; thus, the model learns a mapping 
between the automotive geometry and the drag coefficient.

The primary issue with constructing data-driven models in this application
space is the paucity of freely-available, realistic automotive geometries,
owing primarily to the proprietary nature of commercial designs. A compounding
factor is the nature of the automotive design process, which is driven by 
not just aerodynamic efficiency in mind but with other (and sometimes 
conflicting) considerations such as
aesthetics, regulatory compliance, and manufacturability. (This is in contrast 
with, say, airplane wing design, where (regulatory compliance and 
manufacturability notwithstanding) the design process is far more organic:
one begins with a cross-section of an airfoil with the desired drag and lift
characteristics, then the wing is extruded in the third dimension based on 
loading characteristics.) In practice, the design problem for the aerodynamicist
involves limited scope; small variations in a discrete number of parameters. 
As an example, the aerodynamicist will never be able to redesign the Ford Mustang
to have the same drag coefficient as, say, a Tesla Model S (which is one of
the only passenger cars whose exterior design was driven primarily by 
aerodynamics \cite{tesla}), because a Ford 
Mustang's overall shape and first order drag characteristics are dictacted by 
the need to look like a Ford Mustang, for aesthetic appeal.
This design paradigm makes it all the more important to have a 
rich and variegated training set, because it is the details on a scale much
smaller than the characteristic length scale of the vehicle that one has the
freedom to modify with the aim of improving aerodynamics. In the absence of a
sufficiently large and diverse training set,
models suffer in accuracy and predictive capability. 
In this
work, we develop a methodology of generating geometries for training in a 
controlled manner, starting with a small number of starting cases. By 
generating a dataset with sufficient granularity, we show that data-driven
model predictions show excellent predictive capabilities for quantities of 
interest at different levels of abstraction -- scalar (the drag coefficient), 
vector (the surface pressure along the axis of symmetry), and tensor (the 
surface pressure on the wetted area).

\subsection{Geometry parameterization and dataset generation}
There have been attempts in the literature to address the issue of dataset
generation in aerodynamics problems. Viquerat \textit{et al.} \cite{2dair}
studied the problem of 2D airfoils in the laminar ($Re \ 10$) regime, and 
used
a random process to generate B\'ezier curves with which airfoils were developed.
Their approach --- which used an immersed boundary solver --- showed the 
importance of having sufficienty variety in the training set, but is not 
directly translatable to the 3D, fully-turbulent automotive design problem
for reasons described earlier, as one does not typically encounter realistic
designs through a randomized morphing process. Song
\textit{et al.} \cite{shape} used the versatile ShapeNet database, which is a large 
repository of freely-available CAD models organized hierarchically, in their
investigations of automotive drag. The 
database includes several thousand automobile models of varying level of detail
and quality. They then trained CNNs with planar images of the inputs to predict
the drag coefficient. However, owing to the uncontrolled nature of the dataset,
the model had limited predictive success. The use of a database like ShapeNet,
while appealing from the perspective of quantity and variety of data, 
introduces limitations
simply by the models not being from sources connected with automotive design
and simulation. 
In the complex realm of automotive design, there is a lack of clarity on
the additional coverage in the feature space provided when a new geometry is
considered, owing to the nominally high-dimensional parameter space that the
designs live in. This also leads to confusion over whether predictions made
by models trained on these datasets are actually interpolations or 
extrapolations.
Garcia-Fernandez \textit{et al.} \cite{bus} studied simplified
representations of heavy vehicles, using CNNs to predict aerodynamic quantities
of interest. They followed a similar approach to the other works above, with
a manual parameterization of the Ahmed body to generate a training set. Other
parameterizations explored in the literature is the idea of using a PolyCube
map to represent the geometry \cite{poly}, or using parameterized curves to 
generated two-dimensional silhouettes of cars \cite{guni}, both of which require
manual parameterization, and have been tested on relatively simple shapes.

Virtually all the existing work in the literature uses RANS (Reynolds-Averaged
Navier-Stokes) simulation data as the ground truth to train the data-driven
models. While RANS simulations have been the workhorse for engineering CFD,
it has two primary limitations: the extreme mesh requirements for 
grid-independent solutions in complex geometries, and the inability to capture
inherently unsteady phenomena, such as the strong separation seen in the 
wake of a flow past a bluff automotive body. For these reasons, we use
wall-modeled large-eddy simulations (WMLES) to generate the flow solutions.
Recent advances in this modeling approach have made cost of accurate solutions
of first-order mean quantities at cost comparable with RANS simulations in complex
flows of practical relevance \cite{bose}.

\section{Methodology}

In this section, we describe the dataset generation method we propose. The goal
is to go from a handful of starting geometries, and use them to generate 
several more in a controlled manner. This is done by constructing the convex 
hull of the signed distance functions (SDFs) of their surfaces. We use the 
simplest form of interpolation: barycentric, to generate the convex hull. 
Finally, we require that the process be bi-directional; i.e. the format of 
the projections (intermediate geometries) is the same as that of the bases
cases. Each step in this process is described in detail below.

\subsection{Geometry representation}

We begin with a discretized, representation of the geometry, 
in the stereolithographic (STL) format. The unstructured nature of this 
format makes generation of new geometries by interpolation between basis cases 
challenging
due to the absence of consistent mapping of corresponding points or features
However, an STL 
representation is more readily compatible with unstructured CFD solvers, so we 
introduce an intermediate transformation to a structured representation, and
return to the STL format after any interpolation is performed. To
do this, we develop a ray-tracing tool, the working of which is as follows:

\begin{itemize}

    \item \textbf{Structured mesh initialization:} 
    The bounding box $[x_{\min}, x_{\max}] \times [y_{\min}, y_{\max}] \times 
    [z_{\min}, z_{\max}]$ is determined for the STL tessellation $\mathcal{T}$ by 
    evaluating the 
    axis-aligned bounds of all triangles $\{\triangle_i = (\mathbf{v}_1^i, 
    \mathbf{v}_2^i, \mathbf{v}_3^i) \mid i \in \{1, \ldots, M\}\}$, where $M$ 
    is the total number of triangles in the STL tessellation, and $\mathbf{v}_1^i, 
    \mathbf{v}_2^i, \mathbf{v}_3^i \in \mathbb{R}^3$ are the vertices of the 
    $i$-th triangle. The bounding box is discretized into a structured 
    Cartesian grid $\mathbb{G}$ with dimensions $N_x \times N_y \times N_z$, 
    where each $N_i$ is the desired discretization resolution along each axis.
    It is straightforward to apply a localized refinement in discretization
    in regions where finer features need to be captured, but for simplicity,
    we use a uniform discretization resolution in the whole domain in this 
    study.

    \item \textbf{Ray-tracing:} 
    A set of rays $\mathbf{r}(t) = \mathbf{o} + t\mathbf{d}$ is generated 
    along each principal axis (x, y, z), where $\mathbf{o}$ is the ray origin 
    and $\mathbf{d}$ is the unit direction vector aligned with the axis. The 
    ray origins are placed at the centers of the bounding box faces, thus the 
    total number of rays cast is $N_x \times N_y \times N_z$. For each ray 
    $\mathbf{r}(t)$, the algorithm evaluates possible intersections with each triangle 
    $\triangle_i \in \mathcal{T}$. The intersection test 
    utilizes the Möller–Trumbore algorithm \cite{mt} to determine if and where 
    the ray intersects the triangle, producing an intersection point 
    $\mathbf{p}$ when a hit is detected.

    \item \textbf{Intersect attribution:} 
    Once an intersection point $\mathbf{p} = (p_x, p_y, p_z)$ is detected, it 
    is mapped onto the corresponding grid cell in 
    $\mathbb{G}$. The normalized coordinates 
    \[
    \tilde{\mathbf{p}} = \left(\frac{p_x - x_{\min}}{x_{\max} - x_{\min}}, 
    \frac{p_y - y_{\min}}{y_{\max} - y_{\min}}, 
    \frac{p_z - z_{\min}}{z_{\max} - z_{\min}}\right)
    \]
    are scaled to obtain the 
    voxel indices $(i,j,k)$:
    \[
    (i,j,k) = \left\lfloor \tilde{\mathbf{p}} \cdot 
    \left(N_x - 1, N_y - 1, N_z - 1\right) \right\rfloor,
    \]
    where the floor function $\left\lfloor \cdot \right\rfloor$ converts the 
    continuous coordinates to discrete indices. The intersection grids from 
    ray-tracing along all three principal axes are combined to yield a final 
    binary grid $\mathbb{B} \in \{0, 1\}^{N_x \times N_y \times N_z}$, where 
    each cell is defined by:
    \[
    \mathbb{B}_{i,j,k} = \begin{cases} 
        1 & \text{if } \exists  \mathbf{r}(t) \cap \triangle_i \quad \text{s.t. } 
    (i,j,k) = \left\lfloor \frac{\mathbf{p} - \mathbf{x}_{\min}}{\Delta_x} 
    \right\rfloor, \\
        0 & \text{otherwise}.
    \end{cases}
    \]
    Here, $\mathbb{B}_{i,j,k}$ represents whether voxel $(i,j,k)$ in the grid 
    contains part of the geometry (1) or is empty (0). This transformation 
    yields a discrete voxelized representation of the continuous STL geometry 
    $\mathcal{T}$ within the structured grid $\mathbb{G}$.

\end{itemize}

\begin{figure}[h!]
        \centering
        \includegraphics[width=0.49\textwidth]{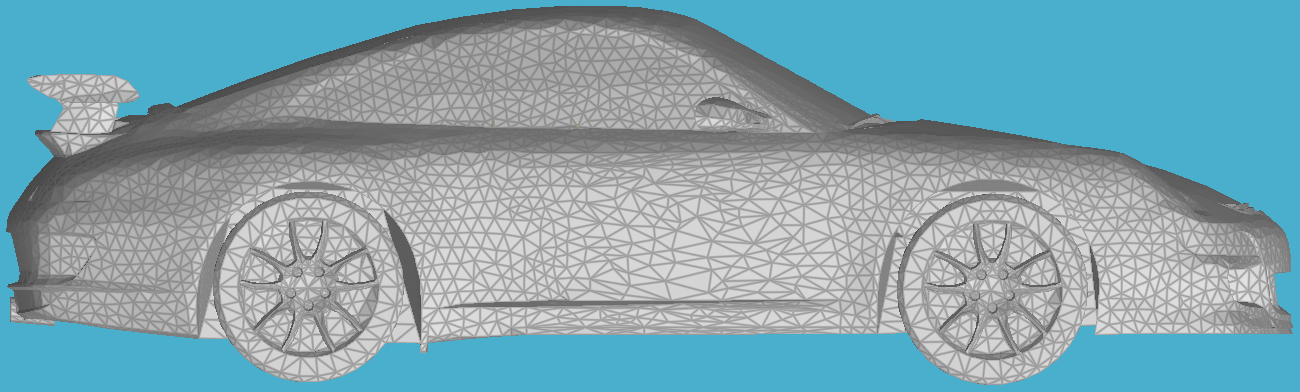}
        \includegraphics[width=0.49\textwidth]{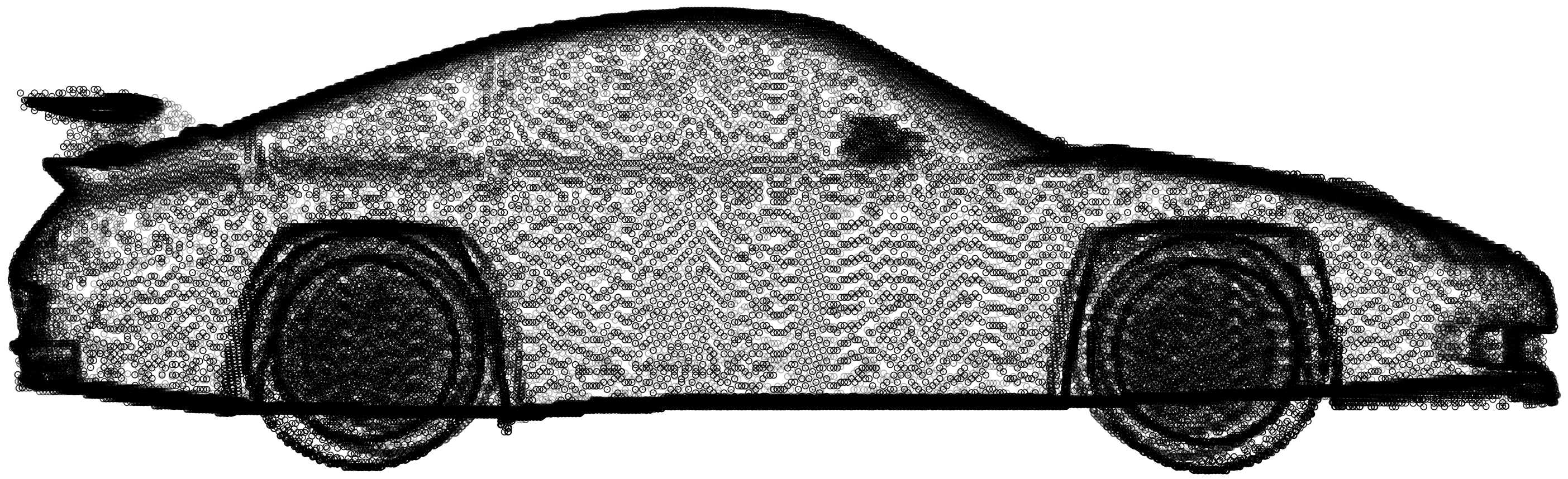}
        \caption{Left: Original STL; right: binary representation}
        \label{f3}
    \end{figure}

Figure \ref{f3} shows an example of the binary representation of an example
geometry of a car, at a ray-tracing resolution of $1024^3$. Every 5th 
intersection is shown. 

\subsection{Geometry interpolation}

With the binary representations constructed as above, we next look at the
question of interpolating between $n$ given geometries, $\mathbb{B}_{i,j,k}^{1},
\mathbb{B}_{i,j,k}^{2}, \dots, \mathbb{B}_{i,j,k}^{n}$.
To accomplish this, we first convert the binary
grids into Signed Distance Functions (SDFs), enabling a continuous
representation of the geometries. For all binary
grids $\mathbb{B}^i_{i,j,k}$, a hole-filling operation is
applied to remove internal
voids that are not connected to the boundary of the grid. 
This involves first dilating the complement of the binary grid
$\overline{\mathbb{B}}^i_{i,j,k} = 1 - \mathbb{B}^i_{i,j,k}$:
\[
\mathcal{D}^n[\overline{\mathbb{B}}^i_{i,j,k}],
\]
where $\mathcal{D}^n$ represents the dilation operation iterated $n$ times.
The hole-filled grid is then the complement of
this dilated result:
\[
\tilde{\mathbb{B}}^i_{i,j,k} = 1 - \mathcal{I}[\mathcal{D}^n[\overline{\mathbb{B}}^i_{i,j,k}]],
\]
where $\mathcal{I}$ represents the invasion step, which identifies the
regions that cannot be reached from the image boundary.
For the filled binary grid $\tilde{\mathbb{B}}^i_{i,j,k}$, the Signed
Distance Function (SDF) $\phi^i_{i,j,k}$ is computed as the difference
between the distance transform of the exterior and interior regions:
\[
\phi^i_{i,j,k} = d(1 - \tilde{\mathbb{B}}^i_{i,j,k}) - d(\tilde{\mathbb{B}}^i_{i,j,k}),
\]
where $d(\cdot)$ represents the Euclidean distance transform. The first
term calculates the distance to the nearest external boundary, and the
second term calculates the distance to the nearest internal boundary. The
SDF $\phi^i_{i,j,k}$ thus yields positive values inside the geometry,
negative values outside, and zero at the surface.
The interpolation involves constructing the convex hull of the basis cases using
a barycentric map over an $n$-simplex. Given the SDFs
$\phi^i_{i,j,k}$ for each geometry $i \in \{1, \dots, n\}$, the interpolated
SDF $\overline{\phi}_{i,j,k}(d)$ is computed as a weighted sum of the individual
SDFs:
\[
\overline{\phi}_{i,j,k}(d) = \sum_{i=1}^{n} w_i \phi^i_{i,j,k}(d),
\]
subject to the constraints
\[
\sum_{i=1}^{n} w_i = 1,
\]
and
\[
w_i \geq 0 \quad \forall \ i \in \{1, \dots, n\}.
\]
Here, the weights $w_i$ are the barycentric coordinates corresponding to the
location $d$ within the $n$-simplex defined by the $n$ input geometries. The
resulting interpolated SDF $\overline{\phi}_{i,j,k}(d)$ provides a continuous
scalar field that represents an interpolation of the original geometries
within the simplex. Though the actual parameter space is high-dimensional,
involving the morphing of a complex, three-dimensional surface, we are able to
effectively parameterize this using the dimension of the simplex $n$.

\subsection{Geometry reconstruction}

After computing the interpolated SDF $\overline{\phi}_{i,j,k}(d)$ using the 
barycentric weights, the interface of the interpolated geometry is extracted. 
The interface corresponds to the numerical zero-level set of 
$\overline{\phi}_{i,j,k}(d)$,
identified by:
\[
\Gamma_{i,j,k} =
\begin{cases}
1 & \text{if } -\epsilon \leq \overline{\phi}_{i,j,k}(d) \leq \epsilon, \\
0 & \text{otherwise},
\end{cases}
\]
where $\epsilon$ is a small positive threshold that depends on the scale of the
models. In our testing we found that
while this value had to be hand-tuned, it was relatively insensitive across
the dataset.

Finally, we reconstruct the surface geometry of the interpolated object. This
is achieved by applying the marching cubes algorithm \cite{mc} to the binary
scalar field derived from $\Gamma_{i,j,k}$. The marching cubes algorithm
generates a triangulated mesh that approximates the surface $\partial \Omega$,
where $\Omega$ is the region inside the geometry corresponding to
$\Gamma_{i,j,k} = 1$.
The vertices $\mathbf{v}_m$ and faces $\mathbf{f}_n$ of the resulting mesh
$\mathcal{M}$ are constructed by interpolating the positions within the voxels
where the interface $\Gamma_{i,j,k}$ intersects the grid. The mesh
$\mathcal{M}$, representing the reconstructed geometry, can be expressed as:
\[
\mathcal{M} = \left\{ \mathbf{v}_m, \mathbf{f}_n \right\} \quad \text{with} \quad \mathbf{v}_m \in \partial \Omega, \quad \mathbf{f}_n \subset \partial \Omega.
\]
Here, $\mathbf{v}_m$ are the vertices positioned on the surface $\partial \Omega$
of the interpolated geometry, and $\mathbf{f}_n$ are the faces connecting these
vertices to form a closed surface.
To refine the mesh and remove artifacts like sharp corners resulting from the
marching cubes algorithm, an iterative Laplacian smoothing operation \cite{yupi}
is applied, 
where each vertex
$\mathbf{v}_m$ is adjusted based on the average position of its neighboring vertices
$\mathbf{v}_m^{(i)}$:
\[
\mathbf{v}_m' = \mathbf{v}_m + \lambda \sum_{i} (\mathbf{v}_m^{(i)} - \mathbf{v}_m),
\]
where $\mathbf{v}_m'$ is the new position of the vertex after smoothing, and
$\lambda$ is a smoothing factor. It is evident that with an increased 
ray-tracing resolution, the reconstruction more closely approximates the 
fidelity of $\mathcal{T}$. 

We have developed a software package that uses the above methodology to 
generate arbitrary-sized datasets. It is available to use on Github: 
\url{https://github.com/benjamark/didymus/}. The ray-tracing has been 
GPU-accelerated and tested on up to 10 billion rays, which provides 
reconstruction at a sufficient resolution for reconstruction of a realistic 
automotive model.

\subsection{Properties of the interpolated SDFs}

The use of the SDFs ${\phi}_{i,j,k}(d)$ for the interpolation provides
advantages for the generating shapes from which a robust reconsturction can 
be made. The properties that make this possible are enumerated below:

\begin{itemize}
    \item \textbf{Continuity:} The interpolated SDF $\overline{\phi}_{i,j,k}(d)$
        is a convex combination of the basis SDFs ${\phi}^i_{i,j,k}(d)$. 
        Given that each ${\phi}^i_{i,j,k}(d)$ is discretely continuous, 
        $\overline{\phi}_{i,j,k}(d)$ is also continuous.
        The continuity argument guarantees that
        for any level $(c)$, the level set $( \{ x : \bar{\phi}_{i,j,i}(x) = c \} )$ will vary
        continuously as $(k)$ changes. The zero level set is not guaranteed
        to exist, but it is \textit{likely} to exist when the basis SDFs
        have sufficient intersection, and are not completely disjoint, which
        is something that can be ensured when positioning the basis STLs.
    \item \textbf{Feature Preservation:} $\overline{\phi}_{i,j,k}(d)$ will 
        not introduce new geometric or topological features 
        that were not present in the basis SDFs ${\phi}^i_{i,j,k}(d)$. 
        Since the interpolation is a convex operation, no new critical points
        or disconnected components are generated. For example, interpolating 
        between two similar car SDFs will not
        introduce a hole or an additional disconnected surface. The zero-level 
        set $\{ x : \bar{\phi}_{i,j,i}(x) = 0 \}$ will reflect the features of 
        the original zero-level sets.
    \item \textbf{Inclusion:} If one basis SDF $\phi^A_{i,j,k}(d)$ is fully 
        contained within another $\phi^B_{i,j,k}(d)$, then the interpolated SDF 
        $\overline{\phi}_{i,j,k}(d)$ will also be contained within $\phi^B_{i,j,k}(d)$ 
        for all $k \in [0, 1]$. This inclusion property ensures that if one shape is 
        nested inside another in the original SDFs, the interpolated shape remains 
        within the boundary of the outer shape. For example, if $\phi^A_{i,j,k}(d)$ 
        is the SDF of a sedan, and 
        $\phi^B_{i,j,k}(d)$ is its estate equivalent, such that the rear roofline
        of the sedan is fully contained within that of the estates, the 
        interpolation will produce shapes contained within the surface of the 
        estate.
\end{itemize}

\section{Simulation and data-driven model design}

\subsection{Geometry and dataset}
\begin{figure}[h!]
        \centering
        \includegraphics[width=0.45\textwidth]{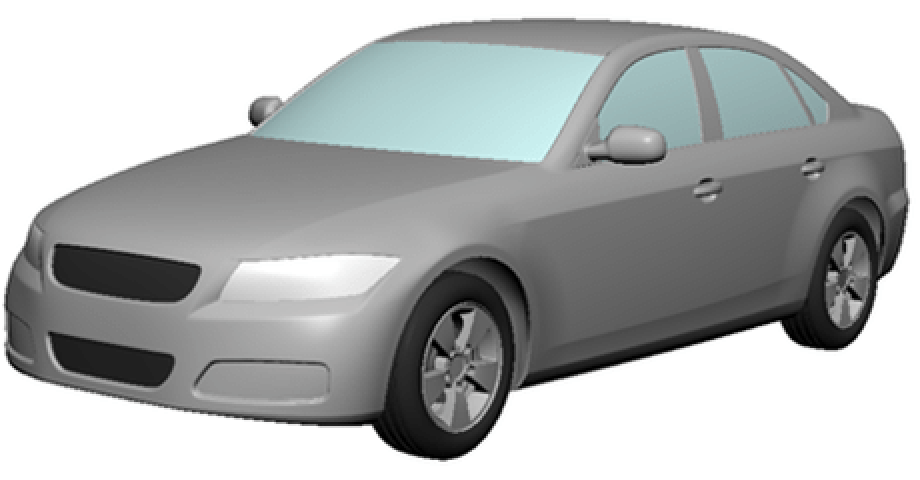}
        \caption{Perspective view of the DrivAer geometry in notchback 
        configuration}
        \label{f1}
    \end{figure}

\begin{figure}[h!]
        \centering
        \includegraphics[width=0.3\textwidth]{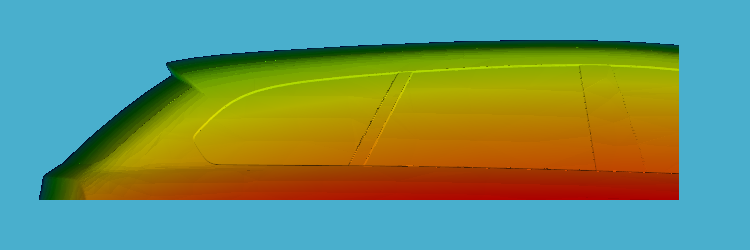}
        \includegraphics[width=0.3\textwidth]{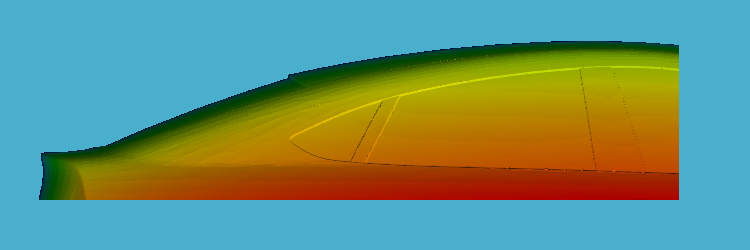}
        \includegraphics[width=0.3\textwidth]{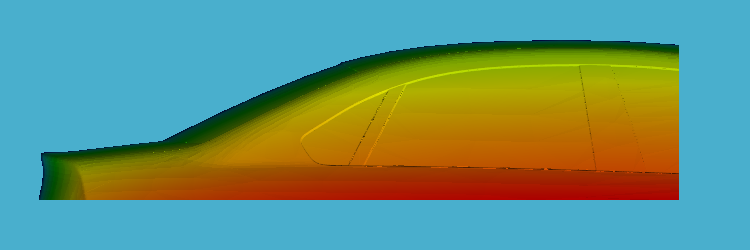}
        \caption{DrivAer back configurations: estateback (left), 
        fastback (center) and notchback (right).}
        \label{f2}
    \end{figure}

To best represent the complexities of a realistic automobile, we consider the
DrivAer model \cite{drivAer}, an open-source midsize passenger
car geometry (Figure \ref{f1}), developed to assess the quality of and 
contrast automotive
aerodynamics investigations carried out using computational fluid dynamics
tools \cite{drivAer2}. The model is available in three configurations: a fastback, estate,
and sedan, shown in Figure \ref{f2}. Elrefaie \textit{et al.} \cite{drivAerNet}
used this geometry as a starting point, but only considered discrete, 
parameteric modifications, and predicted the scalar drag coefficient using 
a point cloud representation. 

While their dataset provides a high quality set of realistic geometries, we 
construct a dataset organically by interpolating between the three DrivAer
configurations to generate a family of designs, our only parameters being the 
choice of basis designs to interpolate between. The three configurations are
different enough so as to provide a sufficient variation in drag coefficient
(the separation aft of the vehicle being significantly impacted by the form
of the rear roof slope),
but also similar enough to provide a controlled testing ground for our
proposed methodology. 

\begin{figure}[h!]
        \centering
        \includegraphics[width=0.49\textwidth]{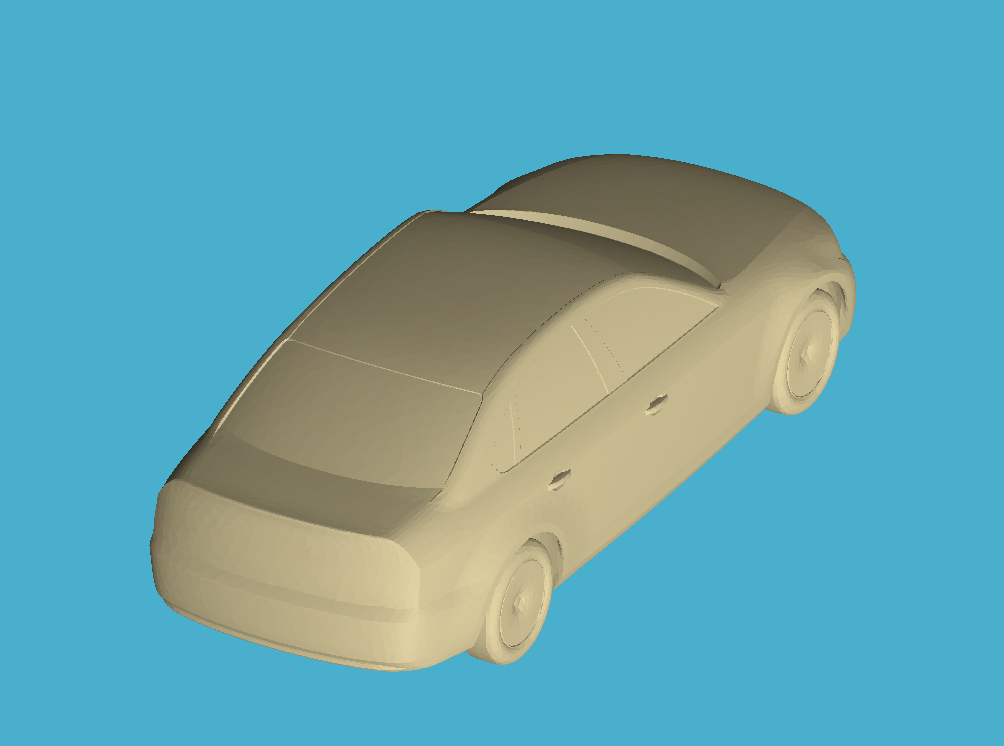}
        \includegraphics[width=0.49\textwidth]{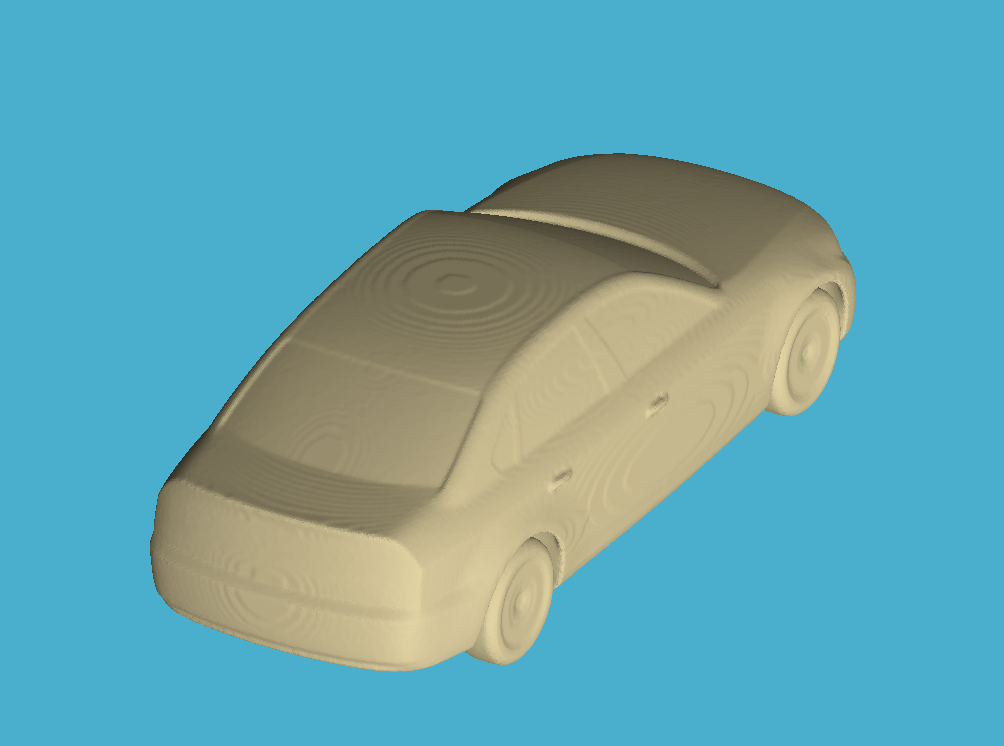}
        \caption{Left: Original STL; right: reconstructed STL.}
        \label{f26}
    \end{figure}

As a test of the ray-tracing approach, we present an example of reconstructing
the notchback model using a ray-tracing resolution of $1024^3$ rays in 
Figure \ref{f26}. We see that
the reconstruction faithfully captures the surface features of the original,
and the new mesh is watertight, which makes generating the CFD mesh simple. We
notice that there is a faint, spurious, low-amplitude rippling in certain regions of low 
curvature and detail, which is an artifact of the finite resolution of the 
ray-tracing. These attenuate with increasing resolution, just as errors in the
STL representation of the underlying CAD curves attenuate with a finer
triangulation. We will show in our discussion that follows that these ripples
do not have a significant impact on the flow solution.

\subsection{Large-eddy simulations}

\begin{figure}
    \centering
    \subfloat[]{\includegraphics[width=0.7\textwidth]{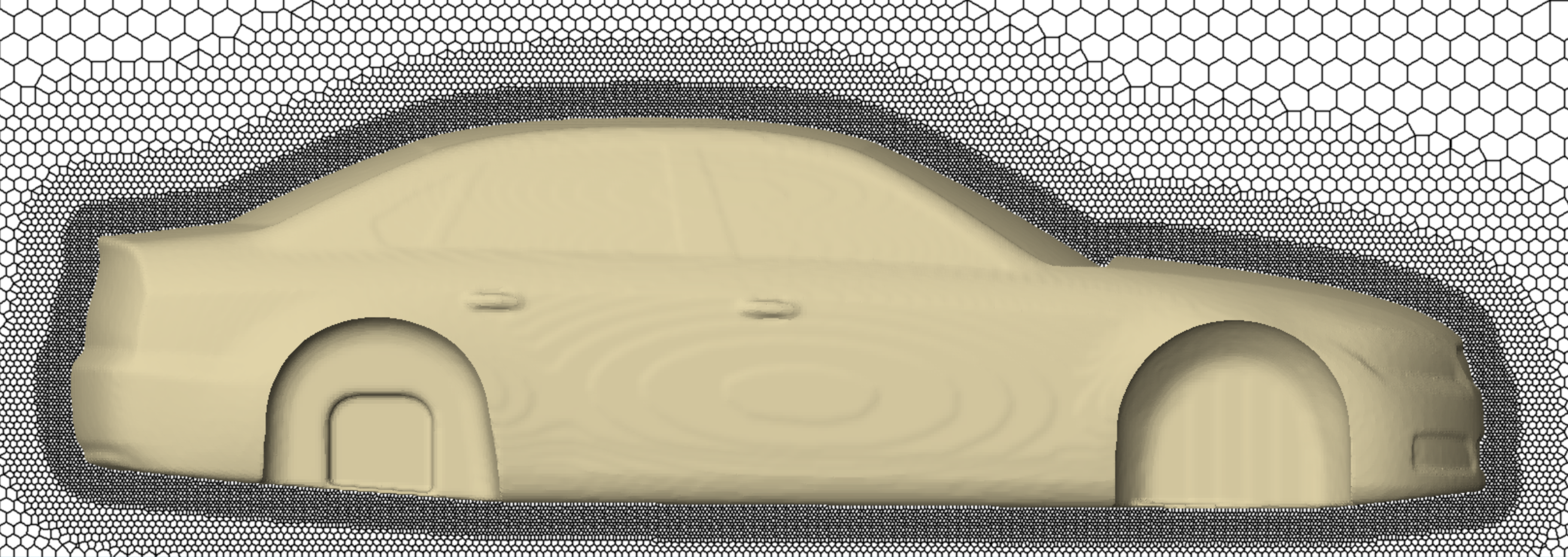}} 
    \\
    \subfloat[]{\includegraphics[width=0.7\textwidth]{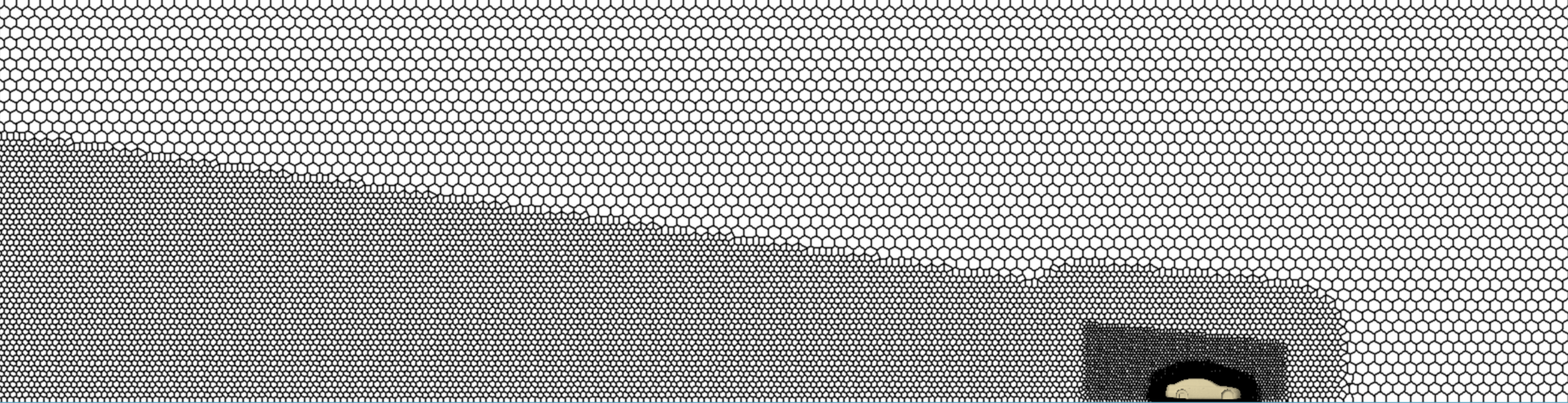}} 
    \\

    \subfloat[]{\includegraphics[width=0.7\textwidth]{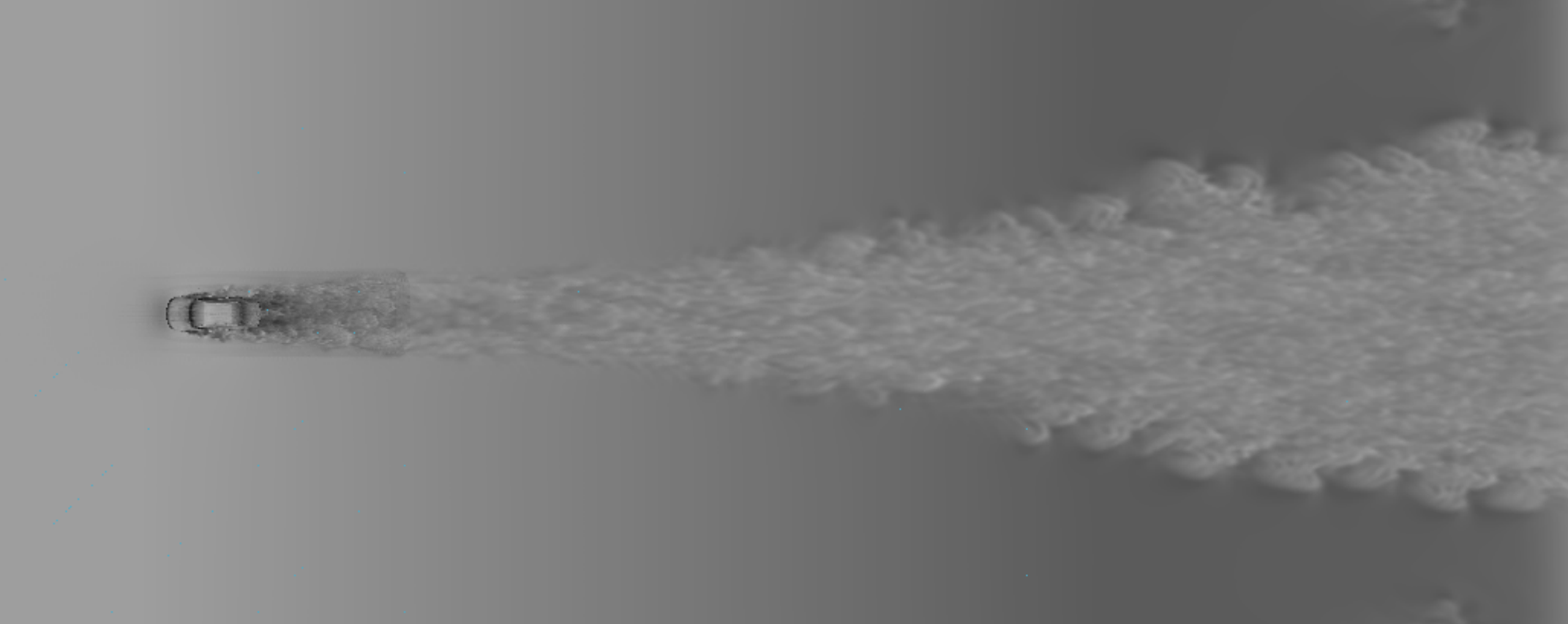}}

    \caption{View of the mesh used for the large-eddy simulations, with
    refinement zones shown: (a) near the car, (b) in the wake. (c) Instantaneous 
    velocity magnitude contours showing wake behind the car.}
    \label{f14}
\end{figure}

\begin{figure}[h!]
        \centering
        \includegraphics[width=0.69\textwidth]{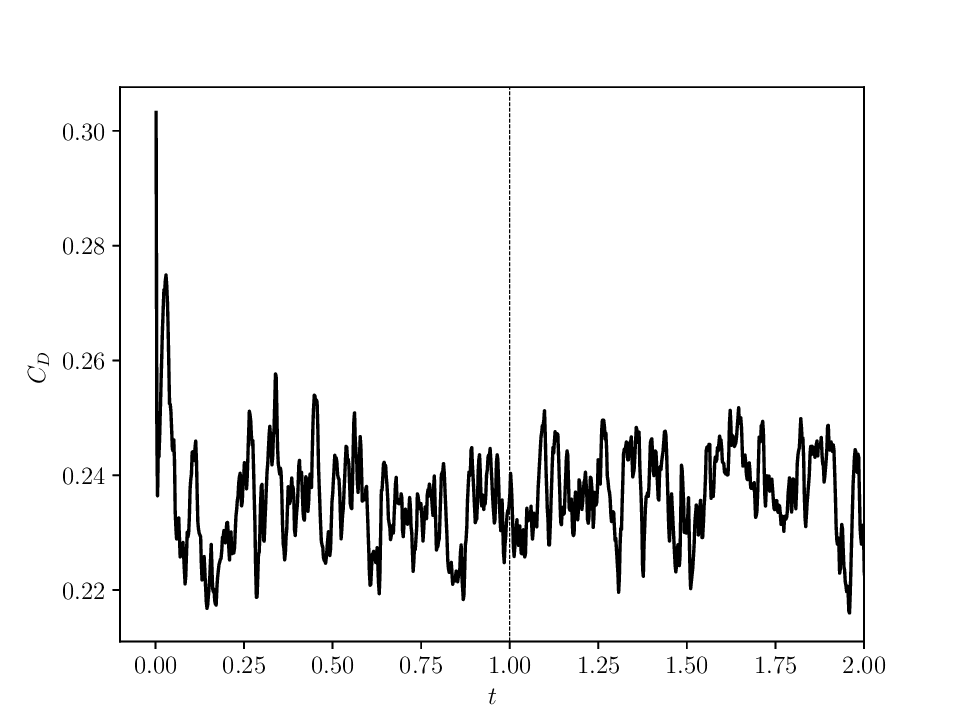}
        \caption{Time history of drag coefficient for a single geometry. The
        dashed line indicates the start of the averaging window. The geometry
        and boundary conditions are to scale, so the time scaling is true to real-
        world time.}
        \label{f25}
    \end{figure}

To generate the training dataset for the neural network, large-eddy simulations 
(LES) of the airflow past the DrivAer geometry at 85 mph with no wheel rotation 
are performed using the low-dissipation CharLES solver. Details of the Voronoi 
diagram–based 
meshing paradigm used in the tool can be found in Brès \textit{et al.} 
\cite{charles1}, 
and the formulation of the low-Mach Helmholtz pressure solver is outlined in 
Ambo \textit{et al.} \cite{charles2}. The code has been validated in aerodynamics 
analysis of a number of high-Reynolds number flow studies \cite{charles3}.
The mesh uses 2 million control volumes, with refinement near the no-slip
surfaces of the car, as shown in Figure \ref{f14}. The resolution on the car
surfaces is $\Delta / L = 0.0023$, where $\Delta$ is the mesh size there,
and $L$ is the characteristic length of the car. The domain is sized at 
$32L \times 5L \times 5L$, with the car placed at $8L$ from the inlet. 
Since the Helmholtz formulation permits low-frequency pressure oscillations,
we use a numerical sponge at $2L$ from the domain boundaries to reduce the time
for the reflecting waves to be damped out.
The Vreman subgrid model is used
to approximate the effects of the subgrid scales on the resolved scales, and
walls are modeled using the algebraic wall model. The flow
is initialized with a converged initial condition from the baseline DrivAer
notchback geometry. The simulation is run for a total time of two seconds, and 
statistics
are collected for the last one second. Figure \ref{f25} shows the instantaneous
drag coefficient, computed as follows:

\[
C_d = \frac{2F_d}{\rho v^2 A}
\]

where $F_d$ is the drag force, $\rho$ is the fluid density,
$v$ is the freestream flow velocity, and $A$ is the frontal area of the car
projected in the direction of the flow. The initial spike is from the recalibration
of the flow to the new boundary conditions (the new geometry), and the averaging
begins well after equilibrium has been established.

The wall time for a single simulation to run to convergence is about 12 minutes
on an NVIDIA H100 GPU \cite{h100}. A sample flow visualization depicting the
velocity magnitude projected on to the floor of the domain is shown in 
Figure \ref{f14}, showing that the 
entire width of the wake is encapsulated in the domain. As mentioned above, 
care is taken to ensure
that the mesh resolution at the wall is comparable to the length scale introduced
by the
ray-tracing, which are $\sim L/1000$, where $L$ is the length of the car,
in order to ensure that discretization 
effects are consistent across different stages of the workflow. The simulation
setup and parameters largely follow our previous work on the same flow \cite{mb}.
The effects 
of the ray-tracing are most evident in the reconstruction in regions of 
low but non-zero surface curvature, and manifest themselves as low-magnitude
ripples, such as in the door panels in Figure \ref{f14} (a).

\subsection{Network I/O}

\begin{figure}[h!]
    \centering
    \includegraphics[width=0.80\textwidth]{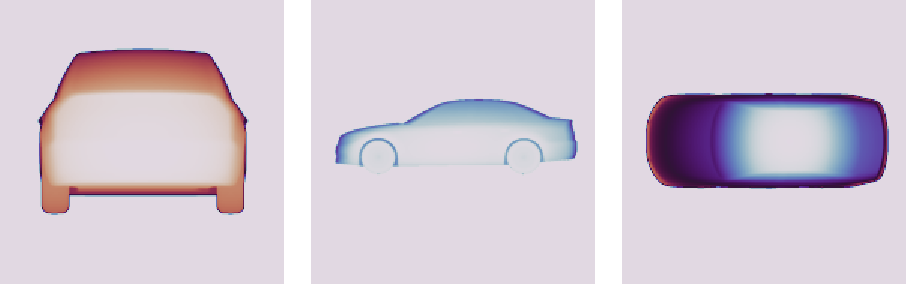}
    \caption{Inputs to networks; left: back view, center: side view, right:
    top view.}
    \label{f5}
\end{figure}

The choice of representation of the geometry is made evident when one considers
that drag depends not on the internal design of the car, but only the wetted
surface, which scales as $\mathcal{O}(N^2)$ on a $\mathcal{O}(N^3)$ volume, 
where $N$ is the characteristic number of points along one dimension of a 
discretized representation of the geometry. Thus, 
projections of the exterior surface are a natural representation choice for the
geometry. For instance, Song et al. \cite{shape} developed a surrogate model that 
used isometric
views of automobile designs from the ShapeNet database \cite{shapenet}.
Projections can be naturally assembled into structured image data that can be
processed by convolutional neural networks (CNNs) \cite{cnn}. CNNs are the 
most popular neural network architecture of choice for image processing, owing
to the shared-weight model of the neurons, allowing for the same feature, say,
an edge or sharp corner in an image, to be learnt across the whole image by
the same kernel. Fewer connections across a single layer of a network leads to
a smaller network that is less prone to overfitting. 
CNNs are also advantageous from an ease-of-training 
standpoint relative to graph networks or point clouds \cite{drivAerNet}, 
the architecture of choice for 
unstructured representations. A graph representation of a geometry is also
highly sensitive to small changes, and not an ideal way to represent a surface
mesh, where the connections between two nodes do not have the same significance
as in an actual graph network; i.e. there can be several valid tesselations of the
same geometry. Our inputs to the model are three planar 
projects of the geometry: the back, the side, and the top (shown in Figure
\ref{f5}), stacked to make three channels. We note here that the this choice
is prompted by the fact that the geometries are identical in the other views,
either due to symmetry, or the fact that the variation is restricted to the
rear roofline. In general, we could use all six views (the current three,
and the left side, bottom, and front views) to get maximum coverage. 
Each pixel records the shortest
distance from the plane of the camera to the point on the vehicle. 
The image resolution is $384 \times
384$, which is chosen to match the resolution of the mesh cells of the CFD
and the resolution of the ray-tracing. This choice is done in keeping with the
need to resolve sufficiently fine changes to the surface. Concretely, taking
the different discretization choices (ray-tracing, mesh, and image resolution)
into account, we expect to detect a change of about 12 millimeters to a surface
in a vehicle that is 4 meters long. The data in the images is binned into 255 bits.

\begin{figure}[h!]
        \centering
        \includegraphics[width=0.95\textwidth]{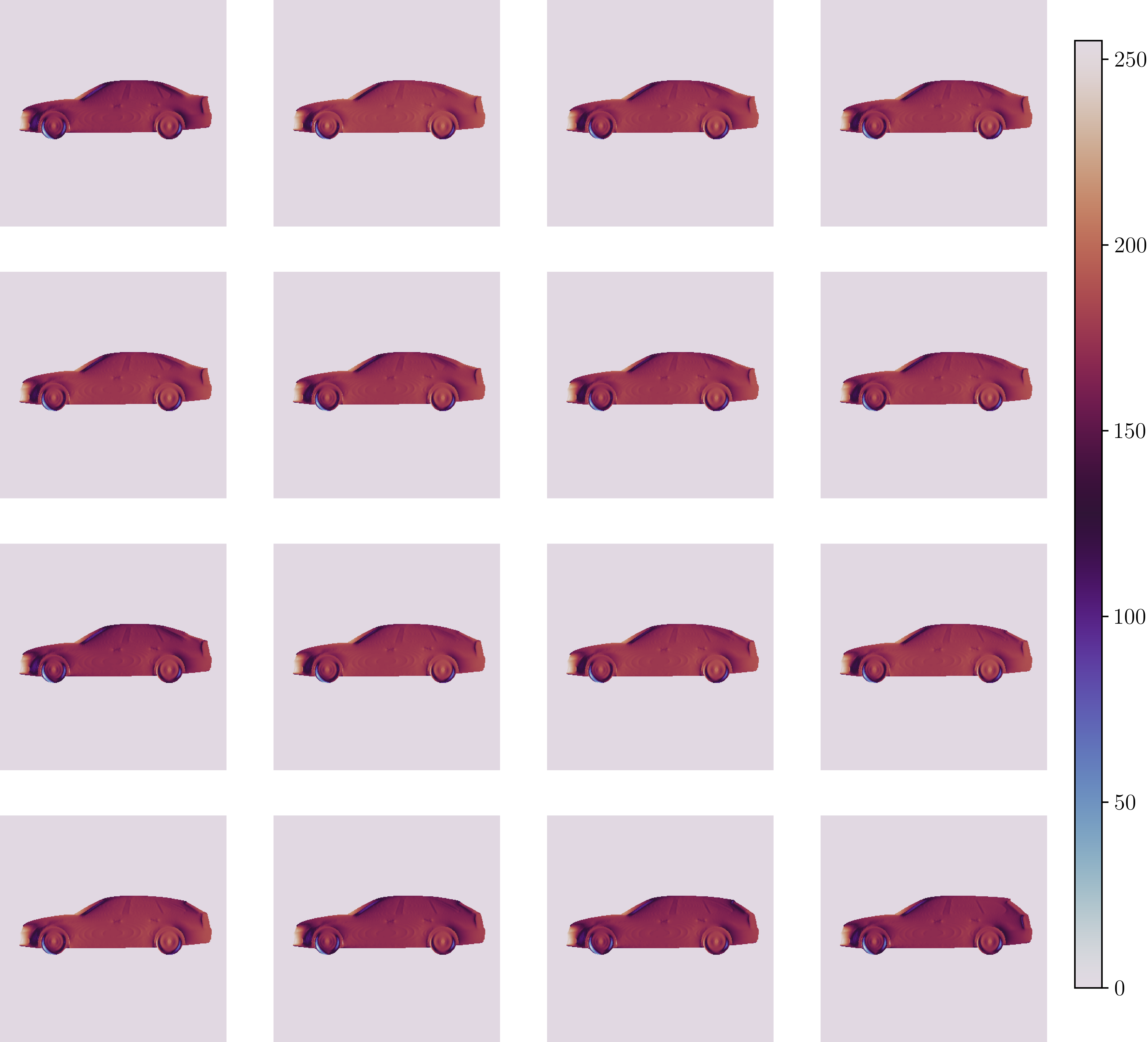}
        \caption{Normalized surface pressures shown on left side views of 
        a selection of samples in dataset, showing a transition from the 
        notchback (top left) to the estateback (bottom right), with the 
    fastback visited along the way.}
        \label{f16}
    \end{figure}

While the drag coefficient is a useful average measure, design is driven by
spatial information of pressure. We thus train networks to predict quantities
of interest at three levels of abstraction: the scalar drag coefficient, the
vector centerline surface pressure along the axis of symmetry of the model,
and the full surface pressure information in the aforementioned three views. 
Individual neural networks are trained for each of these QOI.
Figure \ref{f16} shows a selection of samples from the dataset. One can see
how the roofline is gradually morphed to produce designs intermediate to the 
basis cases. It is also evident how regions of low pressure are more apparent
in the more estate-like designs, which is a result of the larger separation 
bubble that that design tends to produce. With this method of generating
training samples, the parameterization is not tied to discrete design features
such as the door mirrors or handles \cite{mb}, but a comprehensive 
three-dimensional morphological operation, designed to provide a high density 
of data point clustering in feature space.

\subsection{Dataset}

\begin{figure}[h!]
        \centering
        \includegraphics[width=0.90\textwidth]{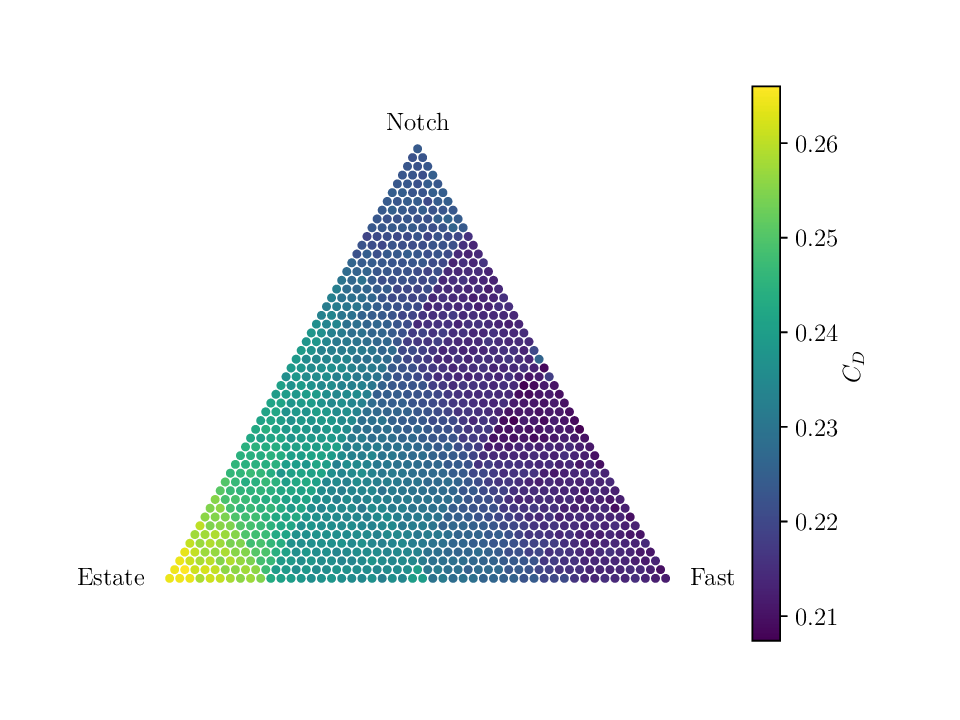}
        \caption{Barycentric map showing drag coefficient of dataset}
        \label{f4}
    \end{figure}

The number of samples per dimension of the simplex is taken to be $n = 50$, 
which gives us $\frac{n(n+1)}{2} = 1275$ total geometries, for which we 
compute flow solutions as described above. 
Since the simplex in 3D is a triangle, it is convenient to plot each sample
in a barycentric map, colored by its drag coefficient (Figure \ref{f4}).
This plot provides an intuitive overview of the entire database, and similarity
of a given sample to a certain corner case is directly available as the 
Euclidean distance between their barycentric coordinates. 
As one expects, the fastback (right vertex) has the 
lowest drag, followed by the notchback (top) and finally the estate (left).
We observe faint banding in the drag coefficient contours, along
directions parallel to the sides of the triangle. This is from stair-stepped
artifacts that stem from the finite resolution used in the ray-tracing, and
reduces as the discretization error for the geometry is reduced.

    \begin{table}[]
        \centering
        \caption{Simulation and dataset parameters}
        \begin{tabular}{ll|ll}
        \hline
        Mesh size       & 2M CVs    & No. of basis cases & 3     \\
        Wall model      & Algebraic & No. of total samples & 1275        \\
        Subgird model   & Vreman    & Normalization   & Min/max  \\
        Simulation time & 2 s       & Train/val split & 1069/141 \\ \hline
        \end{tabular}
    \end{table}

\subsection{Network architecture}

\begin{figure}[!htb]
        \centering
        \includegraphics[width=0.90\textwidth]{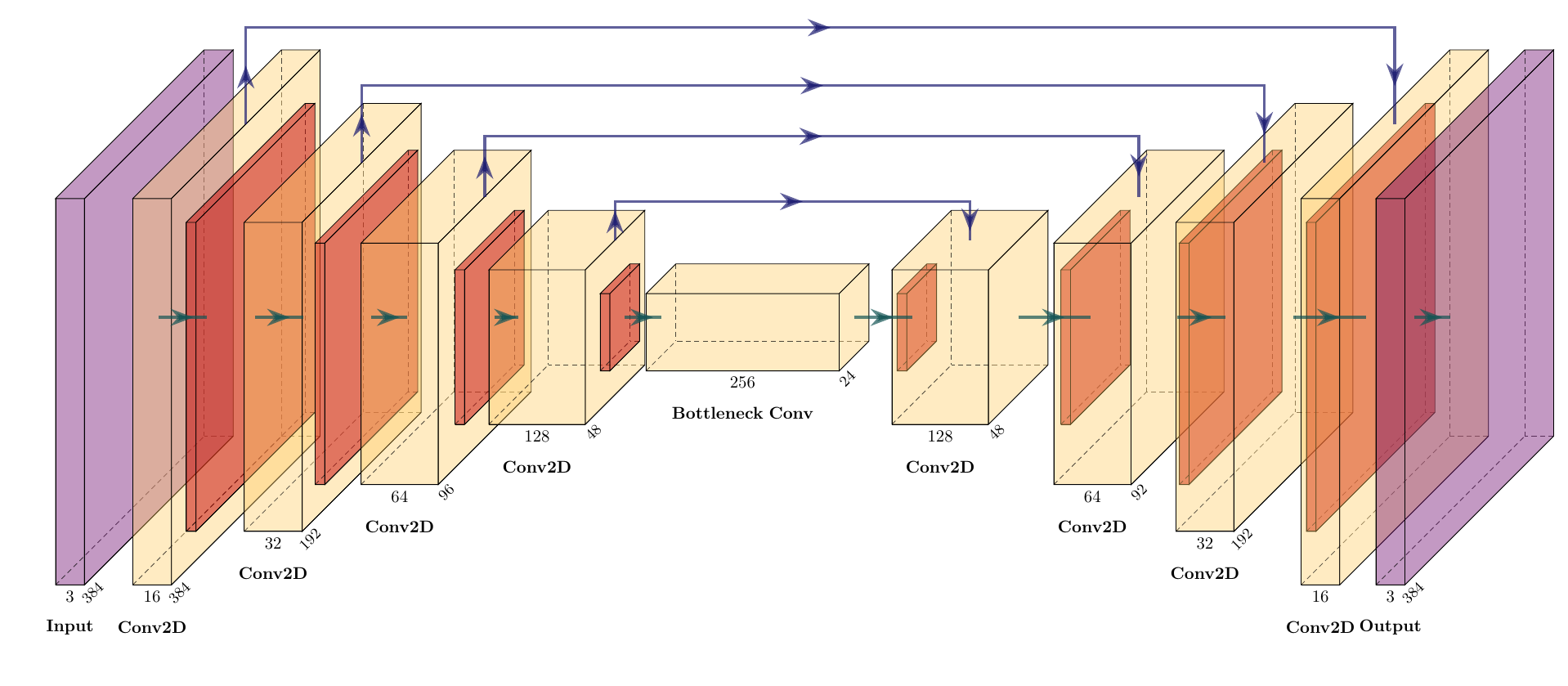}
        \caption{Schematic of U-Net architecture for the tensor model.}
        \label{f6}
    \end{figure}

The architecture of all the networks is convolutional. 
The three networks downsample the input of 
dimension $384 \times 384 \times 3$ down to a bottleneck of $24 \times 24 
\times 256$ in the same way; however, the scalar network has an output of size
1, and the vector network 384, both after going through a fully-connected
layer of size 24. For the tensor case, we use a U-Net architecture \cite{unet}
with symmetric skip connections, as shown in Figure \ref{f6}. The skip
connections serve to greatly improve the recovery of localized information that may
be lost in the several convolutional layers that the information passes 
through \cite{skip}. Maximum pooling layers are used after each convolutional
layer to preserve and highlight important features. A summary of the model
architecture is provided in Table \ref{t1}.
Since the inputs
to all three models are the same images, we keep the same architecture for 
the contracting part of the network, since the feature encoding is expected to
be similar. One may envision this as a form of transfer learning, where the 
scalar model is tuned, and the architecture of the convolutional layers is 
retained for the vector model, and subsequently for the tensor model. All
layers, however, are trained from scratch in the three types of models. 
The hyperparameters for the network --- the learning rate, the
number of filters per layer, and the optimizaer type --- are tuned by a grid 
search.
We perform aleatory and epistemic uncertainty quantification on
the scalar and vector networks by the method of ensembles, and this is 
reported in an appendix.

    \begin{table}[!htb]
        \centering
        \caption{Network architecture}
        \label{t1}
        \begin{tabular}{ll}
        \hline
        Architecture      & CNN/U-Net \\
        Activations       & ReLU  \\
        Pooling           & Max   \\
        Loss              & MSE   \\
        Epochs            & 500   \\
        Optimizer            & Adam   \\
        Learning rate     & $10^{-3}$ \\
        No. of parameters & 50K (1D model), 250K (2D), 1M (3D) \\ \hline
        \end{tabular}
    \end{table}

\section{Results and discussion}

\subsection{Scalar outputs}

\begin{figure}
    \centering
    \includegraphics[width=0.85\textwidth]{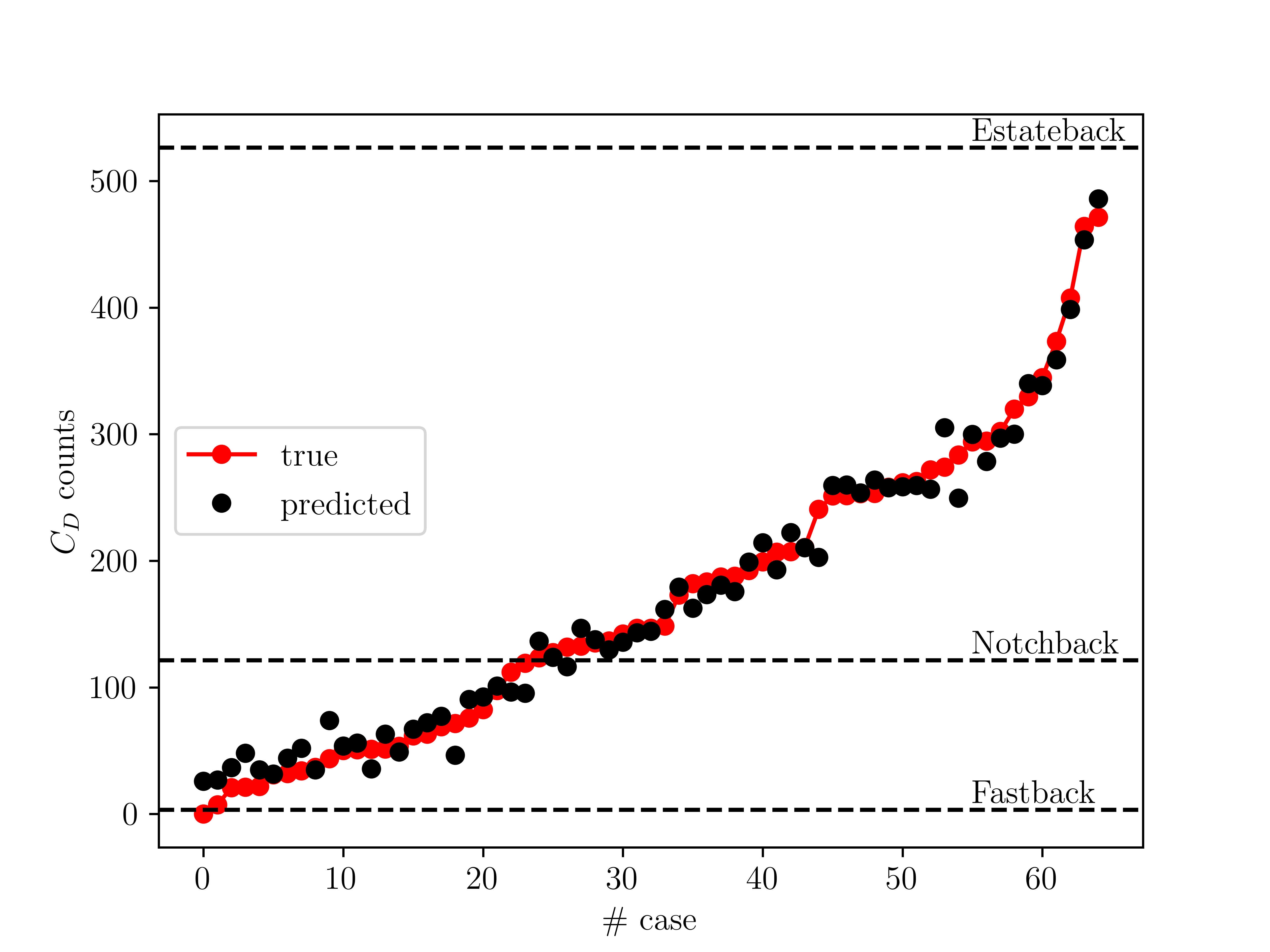}
    \caption{Scalar model performance on test set, organized by increasing
    value of the ground truth. Horizontal dashed lines represent the reference
    drag coefficient values of the basis geometries.}
    \label{f7}
\end{figure}

\begin{figure}
    \centering
    \includegraphics[width=0.69\textwidth]{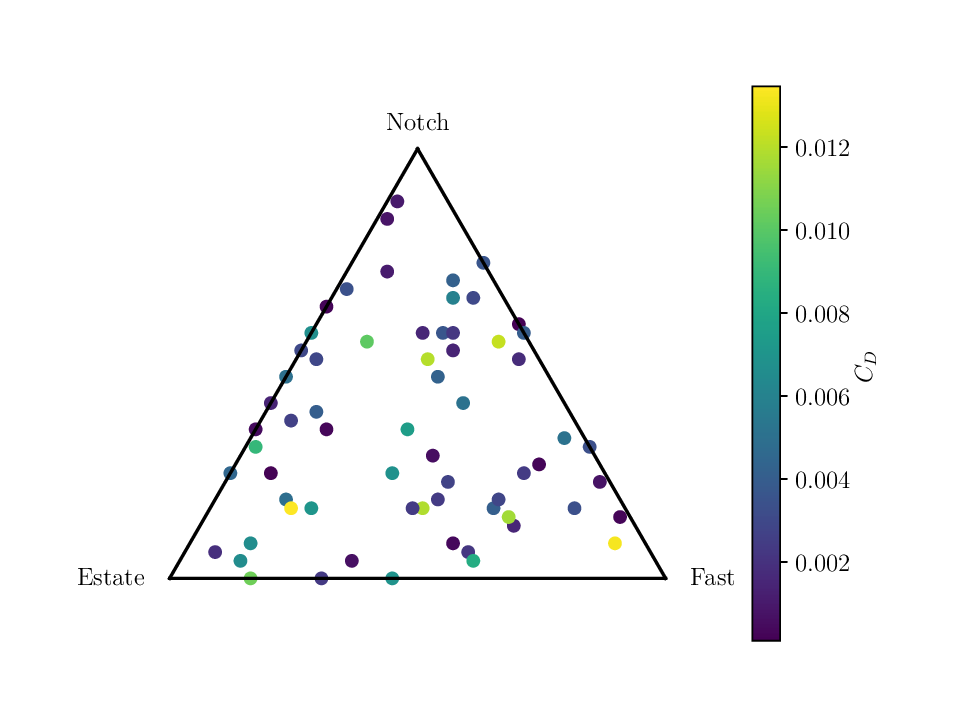}
    \caption{Scalar model error on test set.}
    \label{f23}
\end{figure}

Figure \ref{f7} shows the performance of the model trained to predict the 
scalar drag coefficient on the test set consisting of 65 samples. The figure 
is rescaled by the
lowest value of the ground truth, and shows a range of about 500 drag counts of
variation, defined as
\[
\Delta C_D = C_D \times 10^4
\]
where $C_D$ is the drag coefficient.
The drag count simplifies comparisons across different configurations
by scaling the drag coefficient to a more intuitive range. The measure is 
particularly useful when evaluating subtle changes in aerodynamic
design.
The network shows excellent predictive performance, with a 95\% 
percentile accuracy of 24 counts. The controlled nature of the dataset also
allows us to plot the error on the barycentric map (Figure \ref{f23}), which
gives us insight into the localization of the error. The quantity plotted is
the relative error
$\left|\frac{C_D^{\text{true}} - C_D^{\text{pred}}}{C_D^{\text{true}}}\right|$.
While the current example does not show any error bias toward any of the three
corners, it is evident that in cases where an initial equal sampling of data
points in barycentric space produces an error response surface that contains
regions of high error, a targeted resampling can be done, rather than a blind
and expensive uniform increase in samples.

\subsection{Vector outputs}

\begin{figure}[!htb]
        \centering
        \includegraphics[width=0.55\textwidth]{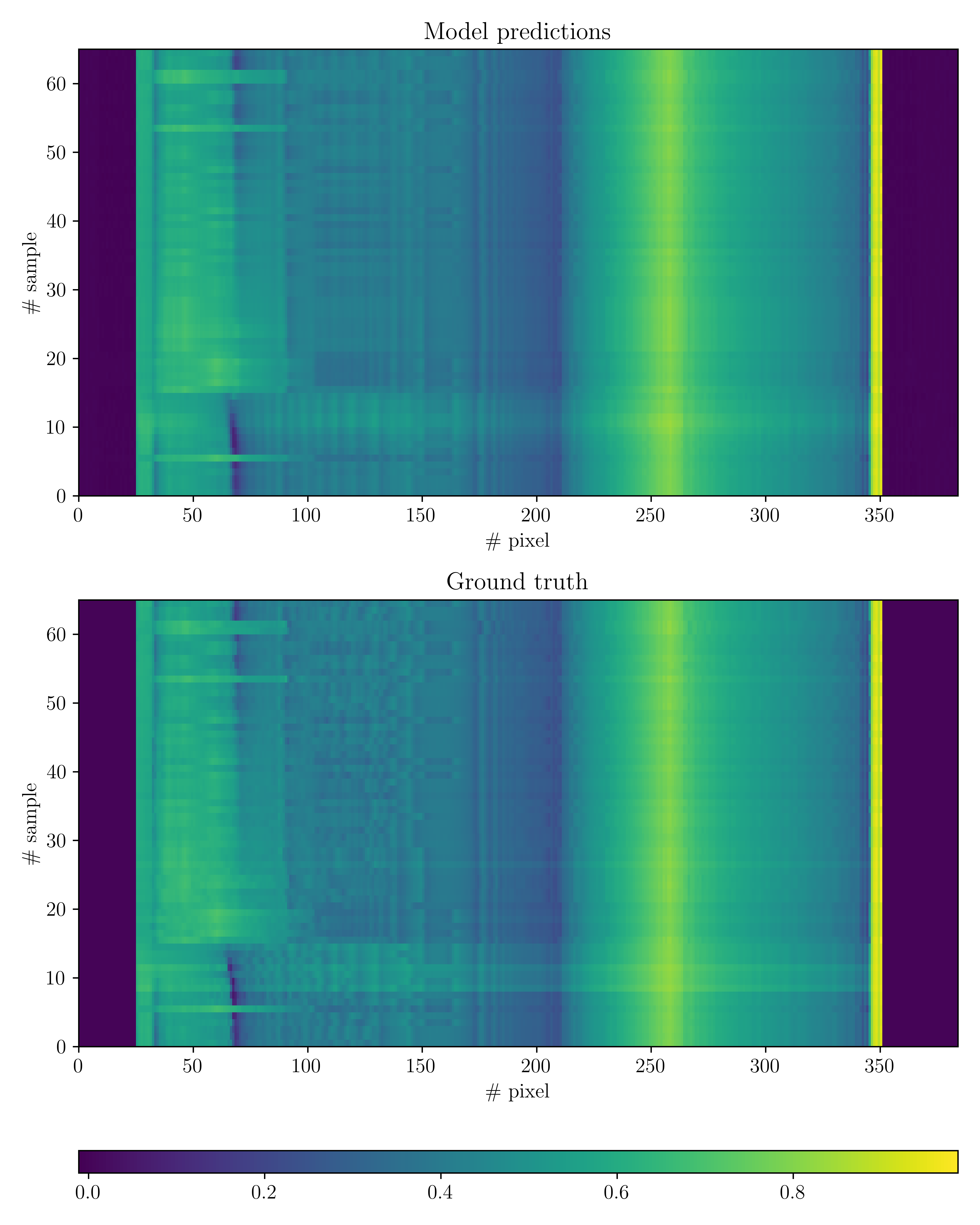}
        \caption{Heatmap of vector model performance on test set. The car is 
        facing right.}
        \label{f8}
    \end{figure}

\begin{figure}[!htb]
        \centering
        \includegraphics[width=0.75\textwidth]{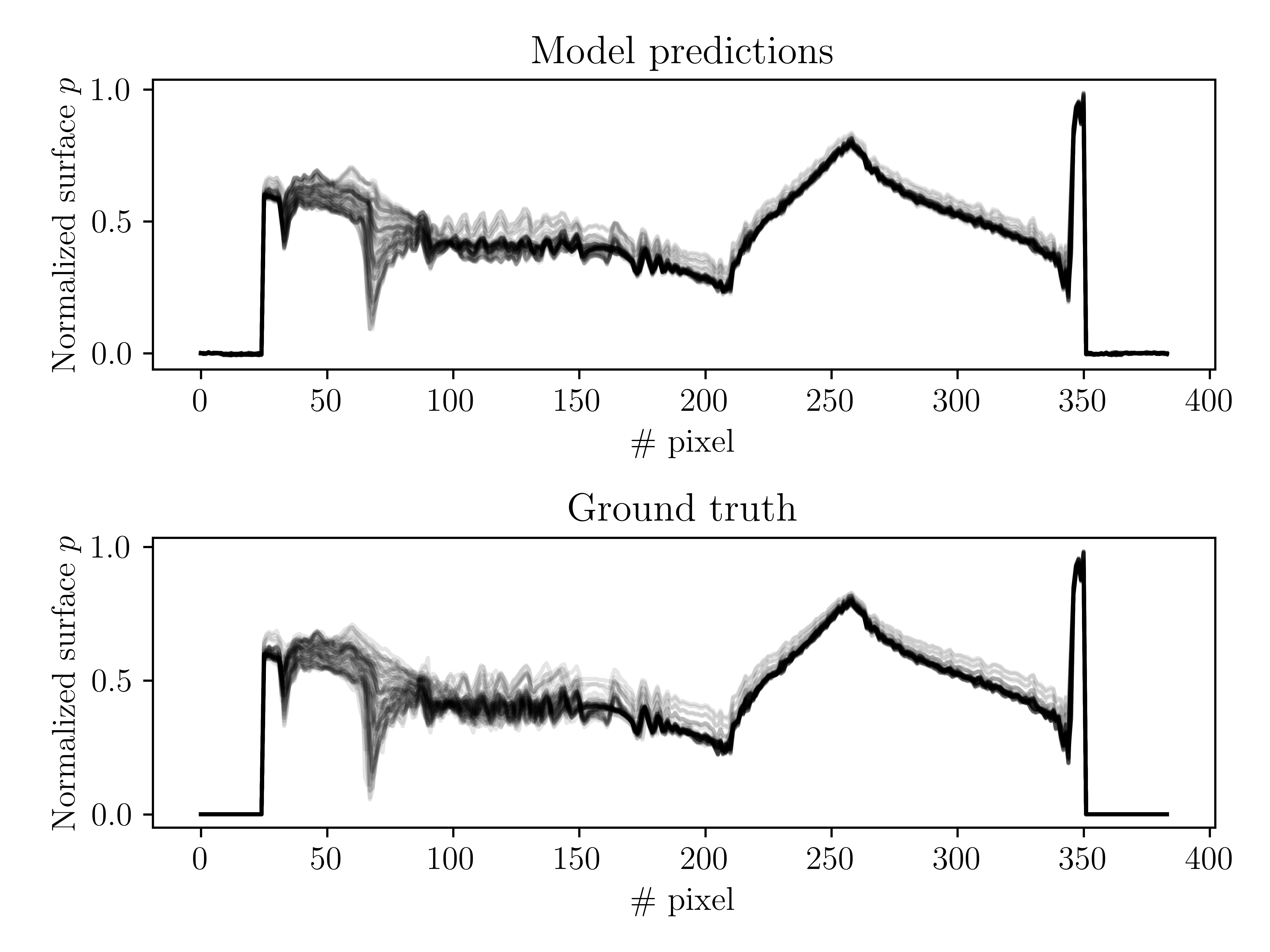}
        \caption{Vector model performance on test set. The car is 
        facing right.}
        \label{f9}
    \end{figure}

We next consider the case of the vector output: the centerline surface 
pressure of the cars, extracted from the same input images. 
Figure \ref{f8} shows a heatmap of the model predictions on the entire test
set, and a comparison with the ground truth shows that the performance is 
likewise excellent in this case. An overlay plot of all the curves in the test
is shown in Figure \ref{f9}, which shows that the spread of 
values between pixel 50 and 150, which represents the zone of high
geometric variance in the dataset corresponding to the back of the vehicle,
is well captured.

\subsection{Tensor outputs}

\begin{figure}[htb!]
        \centering
        \includegraphics[width=0.95\textwidth]{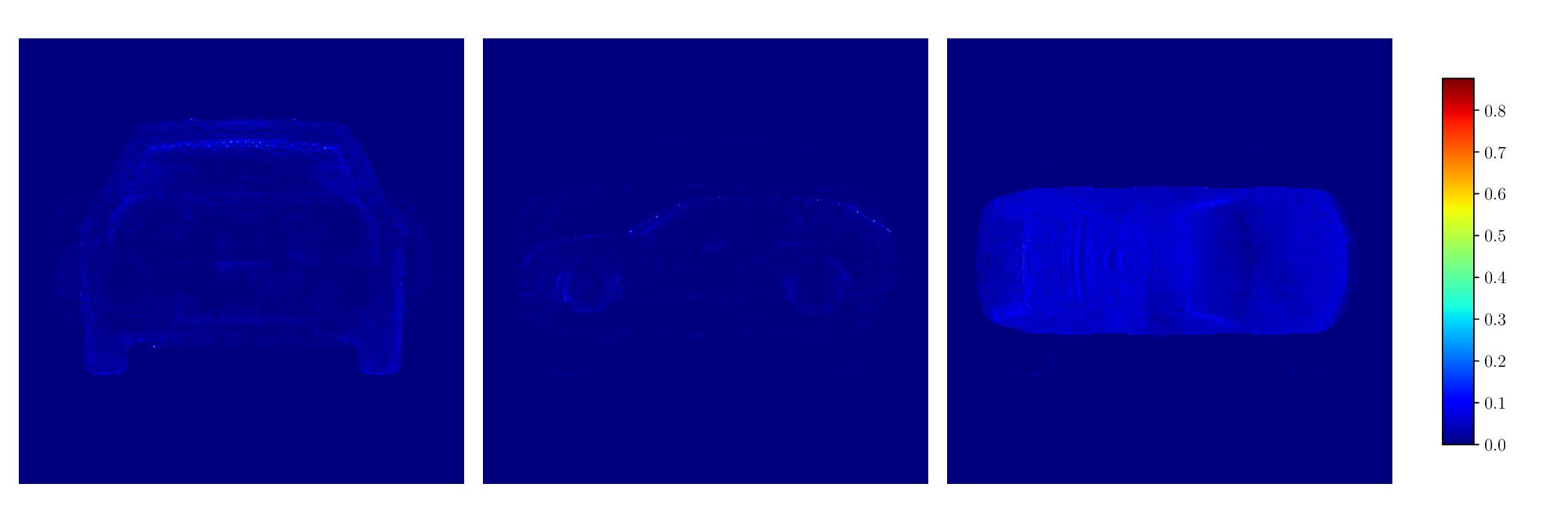}
        \caption{Mean absolute error (MAE) of normalized surface pressure
            over whole testing set. Left: 
        channel 1, center: channel 2, right: channel 3.}
        \label{f17}
    \end{figure}

\begin{figure}[htb!]
        \centering
        \includegraphics[width=0.55\textwidth]{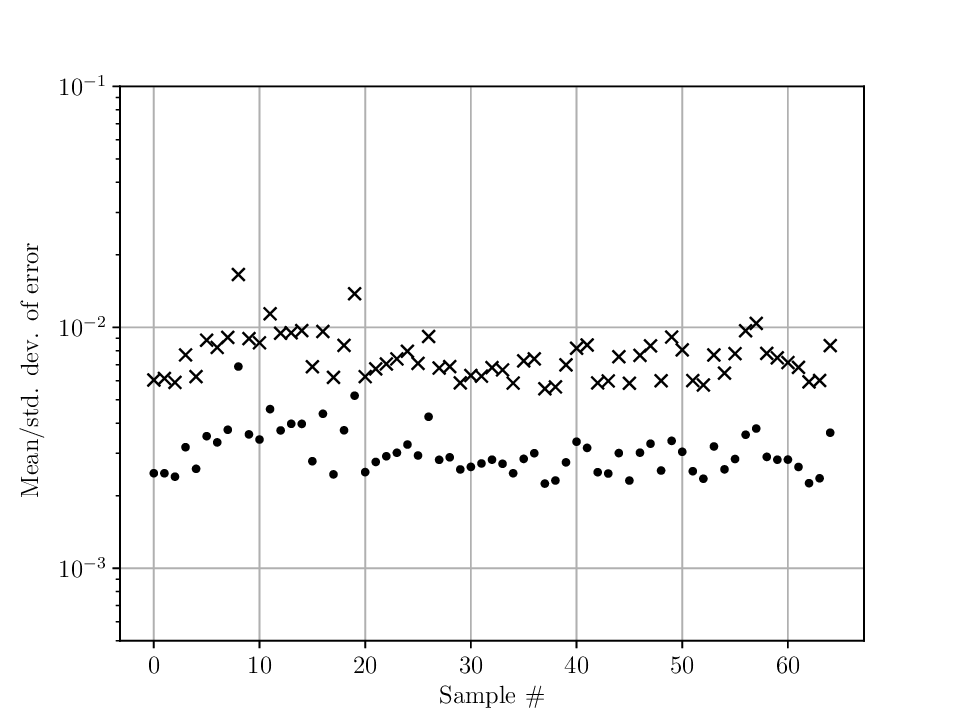}
        \caption{MAE of test set. Mean (dots) and standard deviation (crosses).}
        \label{f18}
    \end{figure}

Figure \ref{f17} shows the average error over the whole testing set of 65
samples for the three different channels. The reconstructions are of good
quality, with precise capture of localized features owing to the skip
connections, which preserve memory across the central constriction. We see
the presence of some crosstalk between the three images, showing that the
three channels are not perfectly disentangled in the training phase. However,
the absolute magnitude of the error is low ($\mathcal{O}(10^{-3}$)). Areas of 
high concentration of the error include the trunk hatch, which is the region
where the flow separates and the impact of the recirculation bubble on the 
surface pressure is seen, and the wheel arches. The mean
and standard deviation of the error over the whole test set is shown in Figure 
\ref{f18};
we see consistent performance across all samples, with a tracking of the two
quantities.

\begin{figure}[htb!]
        \centering
        \includegraphics[width=0.55\textwidth]{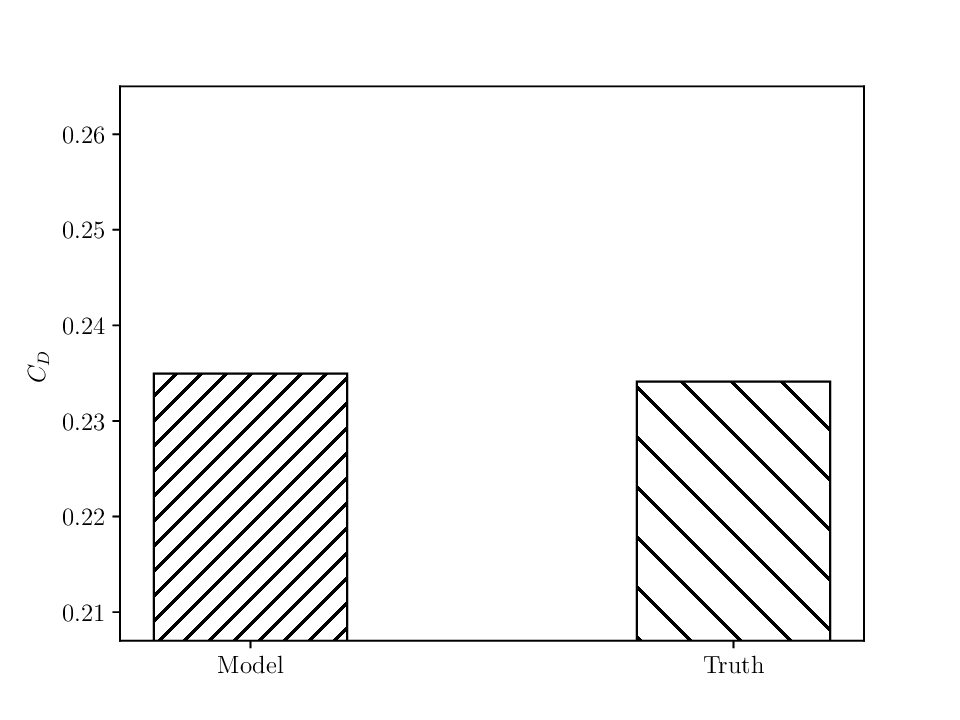}
        \caption{Comparison of scalar level error for one test set sample.}
        \label{f19}
    \end{figure}

\begin{figure}[htb!]
        \centering
        \includegraphics[width=0.55\textwidth]{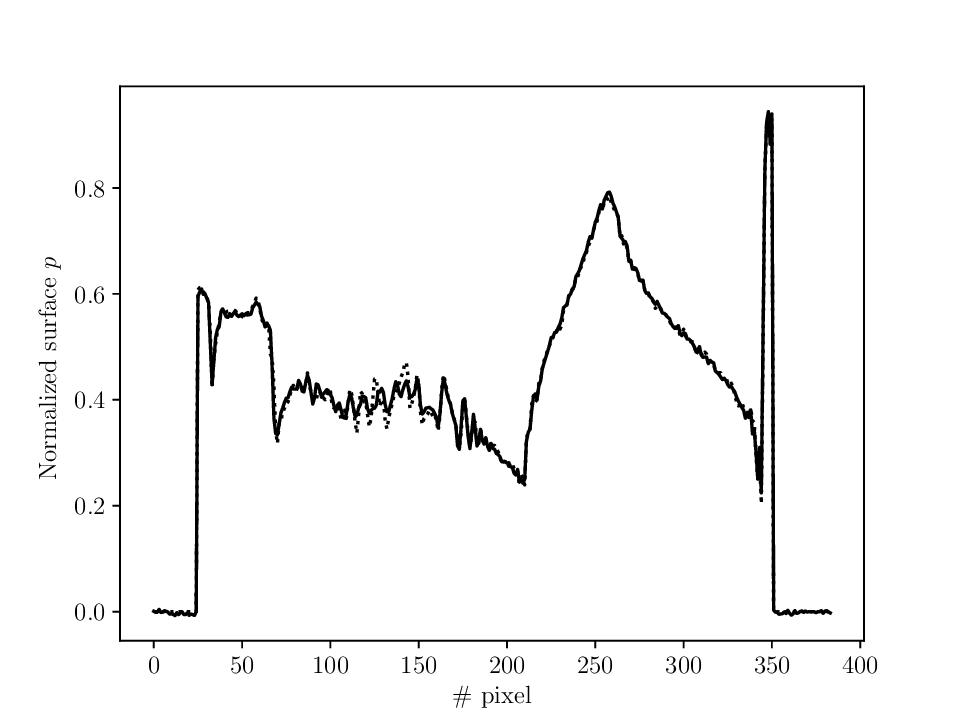}
        \caption{Comparison of vector level error for one test set sample.
        Solid line: model predictions; dashed line: ground truth.}
        \label{f20}
    \end{figure}

\begin{figure}[htb!]
        \centering
        \includegraphics[width=0.95\textwidth]{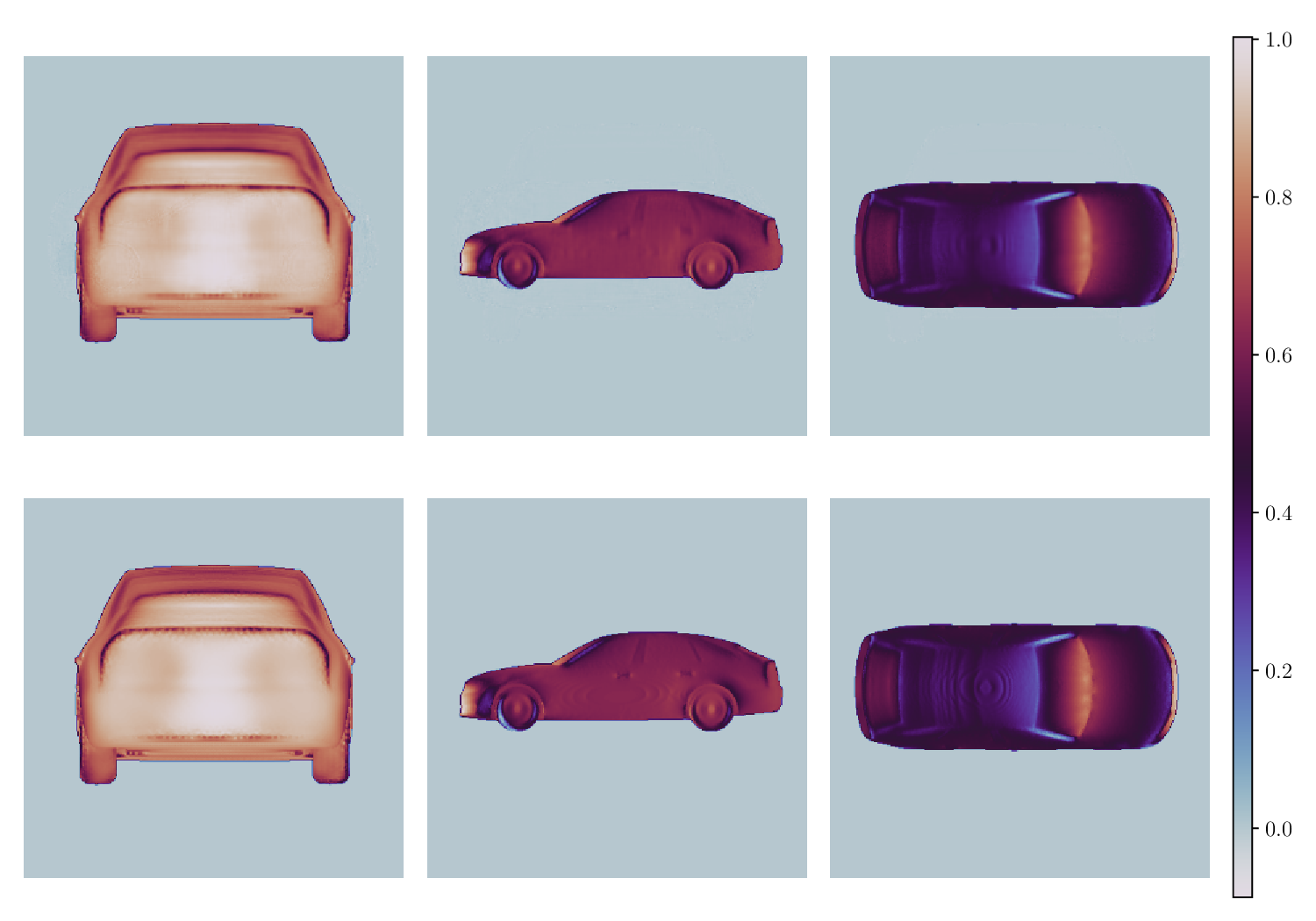}
        \caption{Comparison of tensor level error for one test set sample.
        Top row: model predictions; bottom row: ground truth.}
        \label{f21}
    \end{figure}

We next examine the performance of the networks for a single test set sample
at all three levels of output abstraction. Figure \ref{f19} shows the scalar
QOI; Figure \ref{f20} shows the vector QOI. In both cases we see good 
agreement with the ground truth; we see in the vector case that the only 
significant discrepancy comes from an underprediction of the surface pressure
between pixels 100 and 150, which is the region just aft of the C-pillar of the
car that lies within the separation zone. Figure \ref{f21} shows that the 
tensor case is also reconstructed with good quality,

\subsection{Network extrapolation}

\begin{figure}[htb!]
    \centering
    \includegraphics[width=0.49\textwidth]{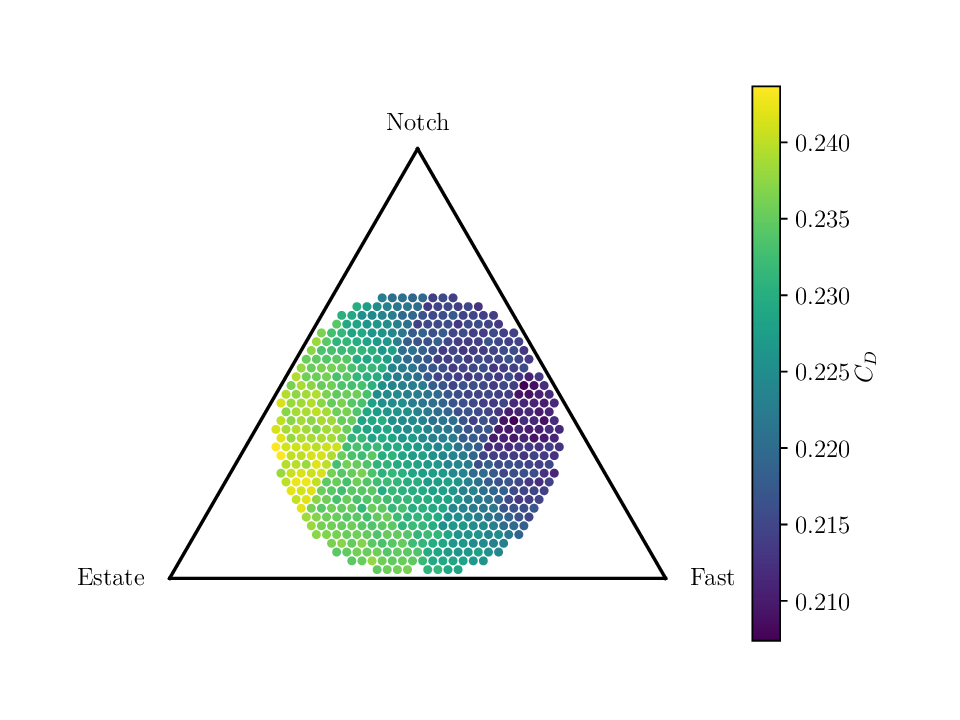}
    \includegraphics[width=0.49\textwidth]{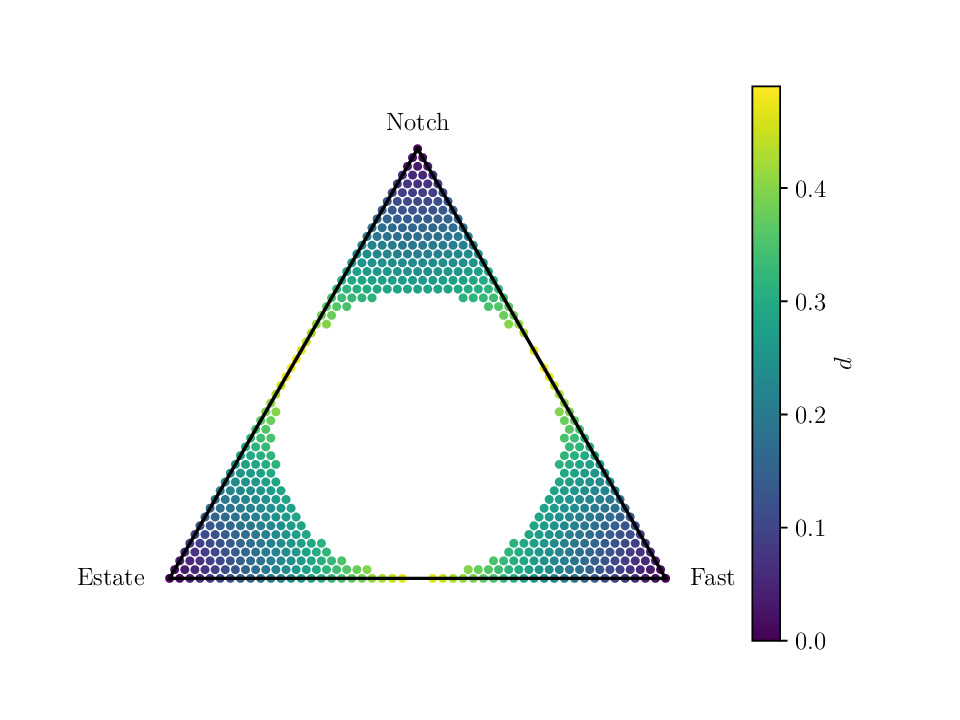}
    \caption{Training set $C_D$ for extrapolation (left) and test set distance 
    from corners (right).}
    \label{f11}
\end{figure}

The benefit of using a controlled dataset is the increased granularity it can
provide from the high-density of samples in the input feature space. This 
attribute also encourages one to consider the possibility of extrapolation to
unseen data. To do this also in a curated manner, we consider the incircle of 
the barycentric map, and construct a training set of points inside it (see
Figure \ref{f11}). Then, we test the model on samples outside the incircle.
This is not entirely undirected extrapolation, since all samples in the 
triangle --- barring points on the edges --- are made up of a linear 
combination of the three basis; nevertheless, this is a case where the 
training and testing distributions undoubtedly differ. Moreover, we have a 
clear visual understanding of which samples in the test set are expected to 
be more difficult to predict, simply by considering the distance from the 
corners (see Figure \ref{f11} (right)). 

\begin{figure}[h!]
        \centering
        \includegraphics[width=0.49\textwidth]{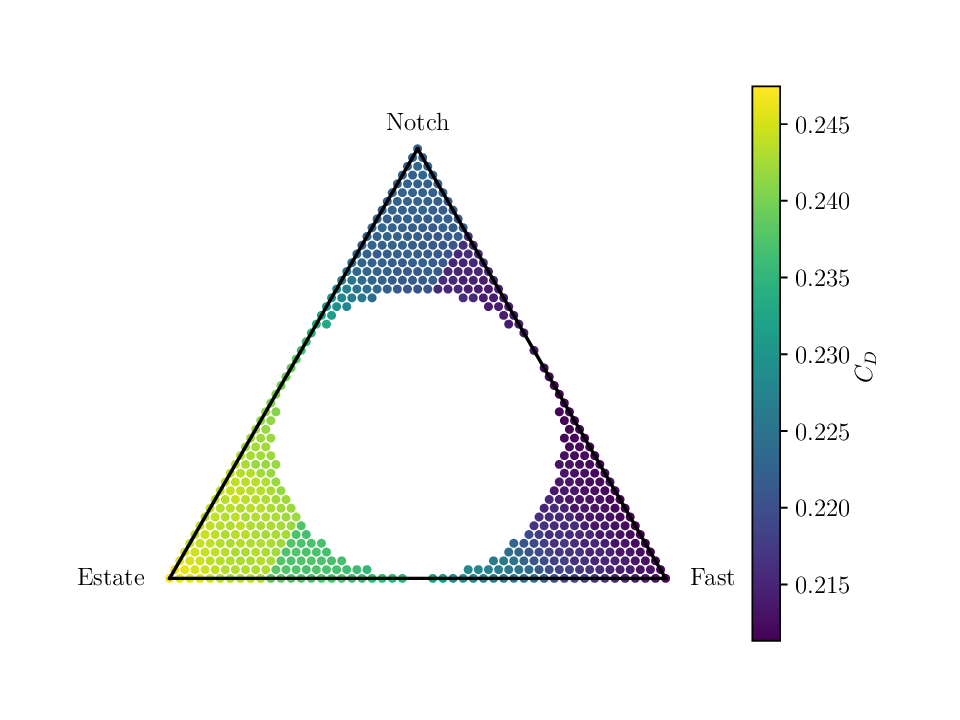}
        \includegraphics[width=0.49\textwidth]{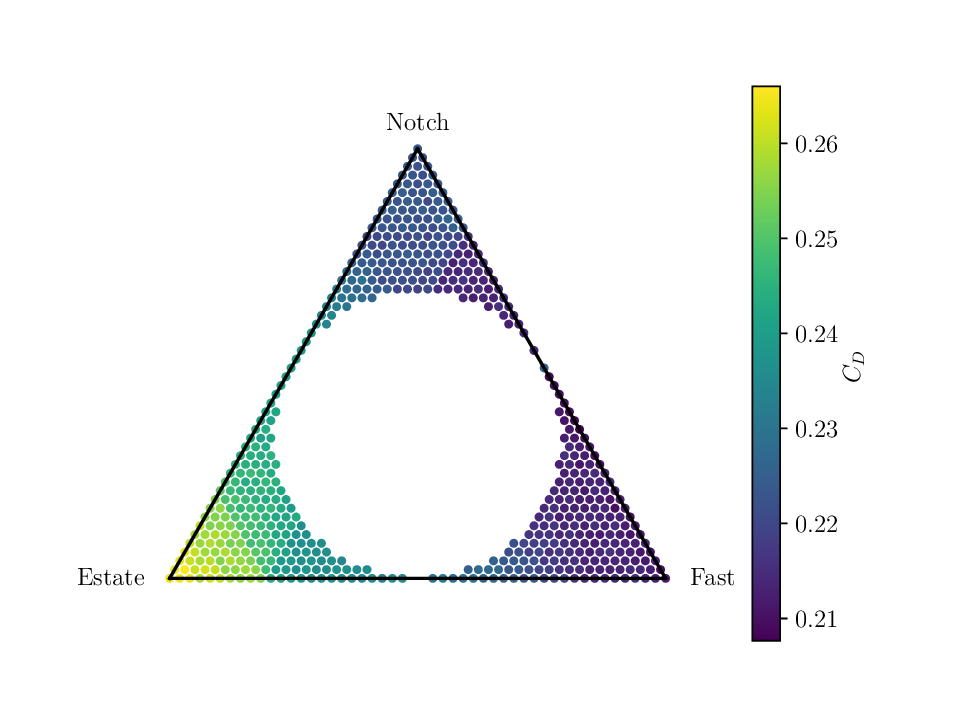}
        \caption{Model predictions (left) and ground truth (right) for
        scalar extrapolation test.}
        \label{f12}
    \end{figure}

\begin{figure}[h!]
        \centering
        \includegraphics[width=0.90\textwidth]{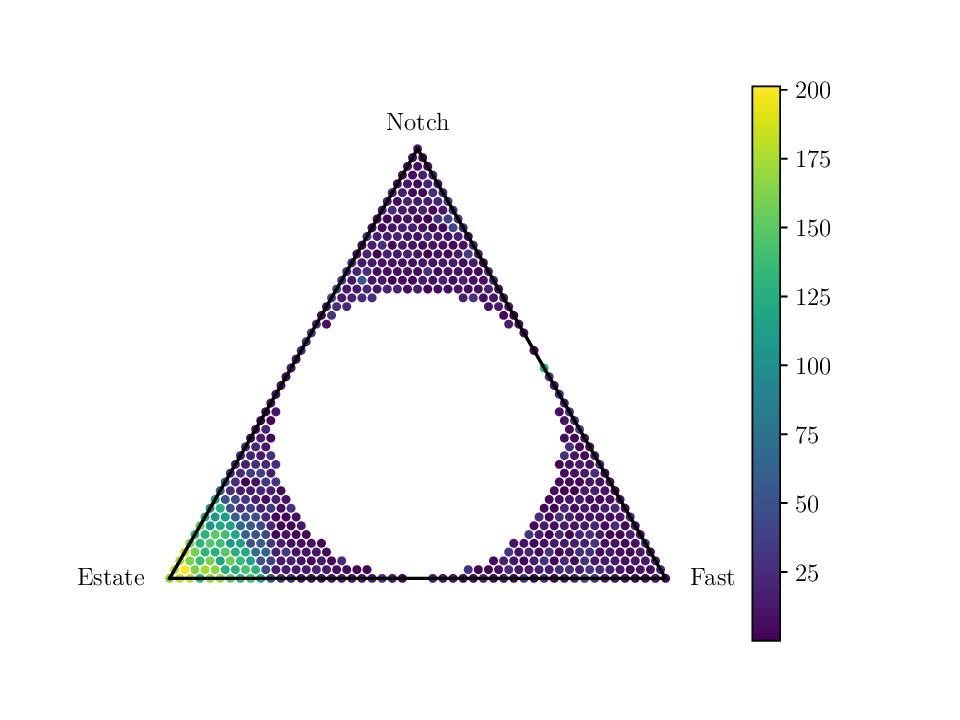}
        \caption{Mean absolute error for model predictions in scalar 
        extrapolation test.}
        \label{f13}
    \end{figure}

Figure \ref{f12} shows the results of these investigations. We see that the 
model does a remarkable job of predicting trends in the drag in the extrapolated
region. However, there is error incurred in the most non-linear portion of the
map, the corner with the estateback, as seen in Figure \ref{f13}. As we 
expect, the error grows with distance orthogonally away from the incircle, 
from the basic 
accuracy of about 25 to 50 counts, to about 150 counts.

\subsection{Effect of uniform barycentric sampling}

\begin{figure}[h!]
        \centering
        \includegraphics[width=0.69\textwidth]{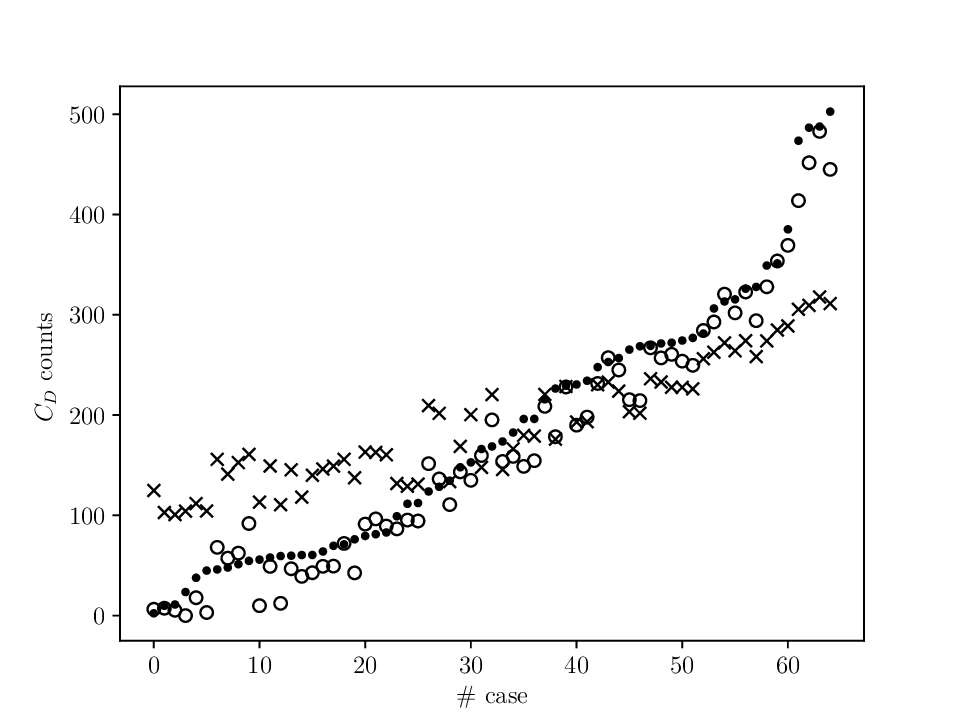}
        \caption{Scalar model performance with reduced dataset sizes; dots:
        ground truth; circles: uniform barycentric sampling; crosses: random
    barycentric sampling.}
        \label{f27}
    \end{figure}

To demonstrate the impact that a controlled sampling has on prediction quality,
we examine a subset of 36 training samples, and train two models: in the first,
the points are sampled uniformly in barycentric space. In other words, the
samples per dimension is 8, which gives a triangle number of 36. In the second
model, the points are sampled randomly in barycentric space, to simulate an
uncontrolled dataset. In this case, we test an ensemble of models with different
random seeds and report a model with representative results. 
The models are trained to predict the scalar drag
coefficient. The results from this experiment are shown in Figure \ref{f27}, on
the same 65 test set samples from earlier experiments.
We see that the effect of the controlled sampling is to produce a network with
a high quality of predictions: the 95\% percentile accuracy is 51 drag counts.
The uniform sampling produces a network with insufficient coverage of the 
different portions of the simplex, and has a much lower accuracy of 108 counts.
This shows that even with a complex and non-linear response function, using
equispaced training points enables better performance.

\begin{figure}[h!]
        \centering
        \includegraphics[width=0.69\textwidth]{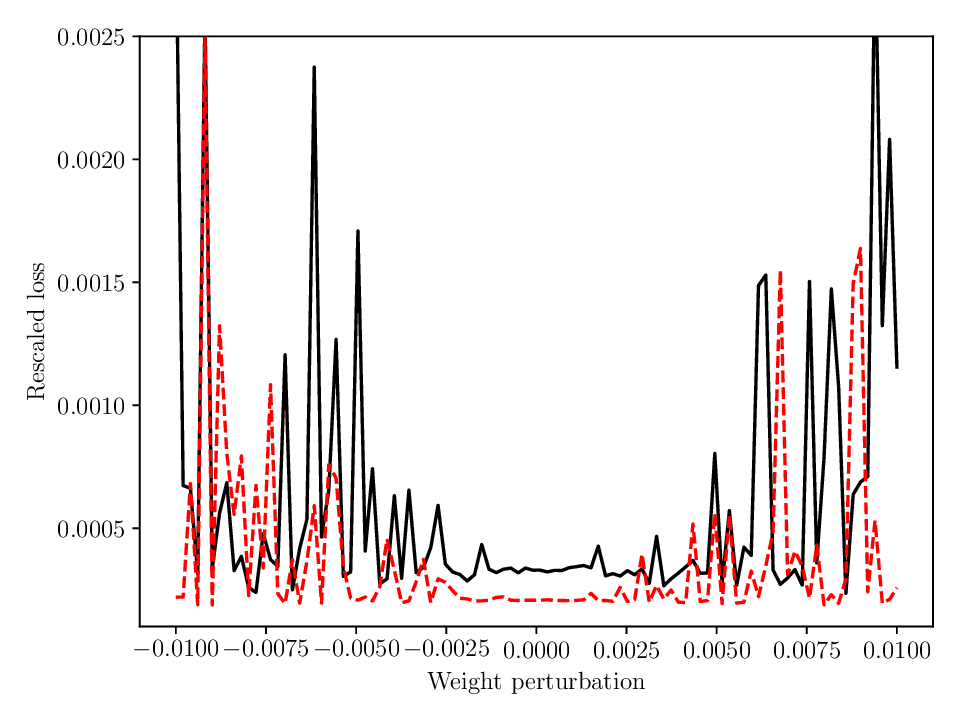}
        \caption{Loss landscape comparison for scalar models; dashed red line:
        uniform barycentric sampling; solid black line: random barycentric 
    sampling. The curves have their means subtracted out to better emphasize
    fluctuations.}
        \label{f28}
    \end{figure}

To understand the difference in performance, we investigate the loss landscape
of the two models, by perturbing the network weights. We apply perturbations
to the weights of different magnitudes, and compute the corresponding change
in loss function. 
This allows us the visualize the stability and generalization
potential of the networks. 
The maximum perturbation strength --- 0.01 --- is chosen
to give an intermediate-scale picture of the loss landscape, between the overall
shape, and the finest details. 
The loss landscapes for both the uniformly sampled 
and randomly sampled models
are shown in Figure \ref{f28}. 
The analysis reveals that the uniform sampling results in a loss landscape with
greater fluctuations, indicative of sharper minima and potentially less
robust generalization. In contrast, the non-uniformly sampled model displays a
smoother landscape, suggesting a broader and more stable basin of minima,
thereby indicating better generalization potential.

\subsection{Comparison with other models: Gaussian Process Regression}

The excellent results obtained by the CNN on the generated dataset prompt us 
to investigate the non-linearity of the problem: i.e. is the prediction of 
aerodynamic drag on this dataset possible with a much simpler model? This 
question is best answered in an \textit{a posteriori} setting, by comparing
the performance of the CNN in the scalar case (the simplest) with another
non-linear model. We avoid the obvious linear model (least-squares regression)
because it is evident from Figure \ref{f4} that the variation in drag 
coefficient over the barycentric map is non-linear. The most suitable 
reference for these investigations, therefore, is Gaussian Process Regression
(GPR). GPR is a
non-parametric, Bayesian inference model that offers a flexible approach to regression
by defining a distribution over functions \cite{gpr}. Since the purpose of this
exercise is to investigate the complexity of the dataset, we directly feed the
model the barycentric coordinates $(x,y)$ that correspond to the geometries. 
This is another benefit to the database generation procedure we have outlined:
each data point is identifiable uniquely by its position in the simplex.

\begin{figure}[htb!]
    \centering
    \includegraphics[width=0.69\textwidth]{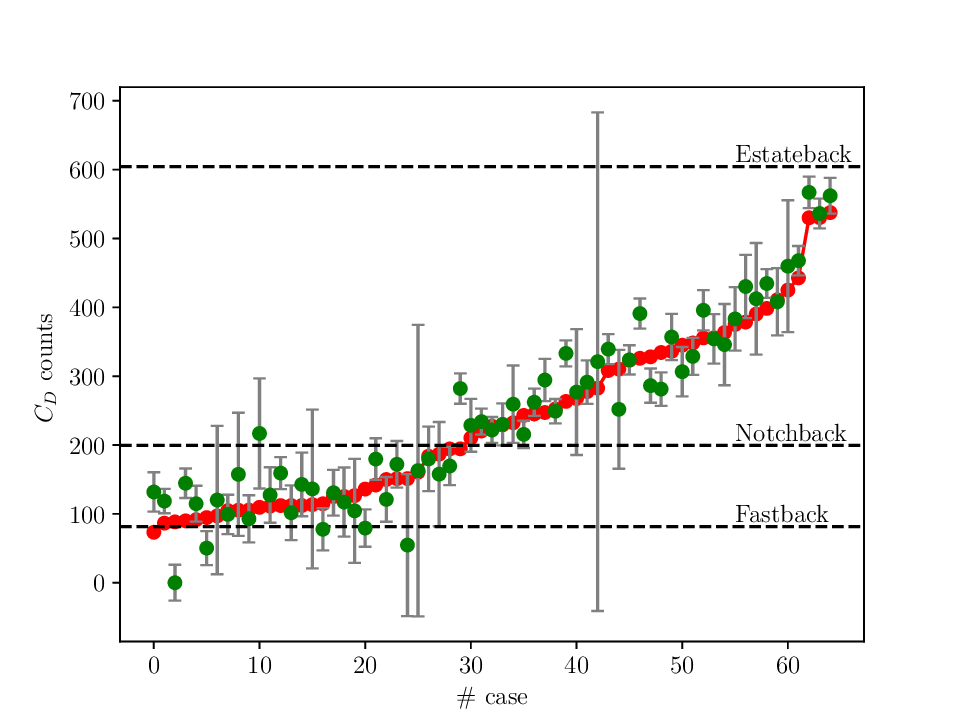}
    \caption{Performance over test set. Red dots: ground truth; green dots:
    results from Gaussian Process Regression. The $y$-axis is zeroed using
    the minimum value of the GPR predictions. Error bars represent 95\% 
    confidence intervals.}
    \label{f24}
\end{figure}

The covariance function (or kernel) plays a critical
role in the performance of GPR. 
We use a composite kernel $k(x_i, x_j) = \sigma_f^2 \exp\left(-\frac{1}{2\ell^2} \|x_i - x_j\|^2 \right)$, 
where $\ell = 1.2$ is the length scale and $\sigma_f^2 = 1.0$ is the signal 
variance. The model was optimized by using the maximum likelihood framework
for hyperparameter tuning, with 10 restarts and a noise term $\alpha = 10^{-10}$.
We observe from Figure \ref{f24} that the tuned GPR model, while predicting 
the test set with fairly good accuracy ($R^2$ score 0.92), still struggles 
to match the performance of the CNNs. While the 95th percentile accuracy for the
CNNs was 24 drag counts, it is 58 for the GPR. We reiterate that the variation of 
$C_D$ across the three designs is primarily due to the separation from the 
roofline, which varies in a complex, three-dimensional way across the dataset.
Thus, GPR requires an even higher sampling than the 50 points per dimension
we have used in this case. Since GPR models the prediction as a distribution,
one can directly obtain uncertainty quantification bounds from the standard
deviation of each output. This is also plotted in Figure \ref{f24}. A comparison
with ensemble uncertainty quantification of the CNNs is shown in an appendix,
and one observes that the confidence intervals for GPR are much broader for
the same dataset.

\section{Conclusions}

In this work, we address the problem of availability of high-quality geometries
for training data-driven models for aerodynamics analysis. Focusing on 
automotive drag prediction, we develop a method by which an arbitrary database
of high-quality, realistic models can be systematically generated from a small
number of starting designs. Our workflow involves the conversion of the 
STL representation of the geometry to an intermediate binary representation 
using an open-source ray-tracing tool we have developed, followed by 
barycentric interpolation to generate new designs. The method has been 
tested using the three starting configurations of the DrivAer model, and we 
show that the performance on quantities of interest of various complexity ---
scalar, vector and tensor --- is excellent. 
Care has been taken to ensure that we are predicting variations in morphology
on a sufficiently fine level, in keeping with practical considerations for 
automotive design.
We attribute the performance of our models to
the granularity of the dataset, which allows the networks to learn a smoother
representation of the drag than would be possible with an unstructured dataset
made up of an assortment of geometries. To show this, we conduct \textit{a
posteriori} analysis of two networks trained on uniform and non-uniform
barycentric sampling of the data respectively, and show that the former has 
a smoother loss landscape, leading to more robust generalization and predictive
accuracy. We believe this method could be 
the basis for a universal drag predictor, given enough training samples in a 
sufficiently high-dimensional space. Ongoing investigations are centered 
around designing a dataset that could potentially work with all cars of a 
certain type, for e.g., sedans, and the task is to then identify what makes
a suitable corner geometry. It is of course evident that if two completely
incompatible shapes are chosen for the corners, interpolation between them 
would result in shapes not likely to be used in practical design; however, the
key is that the network learns about the surface pressure on these shapes
regardless of whether or not the intermediate designs are practical.
Other work in the field has used transfer learning
to good effect; Song \textit{et al.} \cite{shape}, for instance, use a pre-trained ResNet
for their investigations, with only a few trainable layers at the end. It is
possible that as the databases grow in size, the training process could 
benefit from a network that has had a broad exposure to a variety of inputs
and can recognize fundamental image building blocks already. We are also 
interested in exploring other representations for the geometry. Images and 
CNNs have the benefit of ease of training, but suffer from the issue of 
occlusion in non-convex geometries (which includes most automotive shapes), so
alternative representation approaches such as FNOs \cite{fno}, which have been
used in conjunction with graph representations of geometries \cite{sdfs}, could
be used. Our present workflow for generating geometries is easily compatible 
with any of these existing approaches. Another important piece of the problem 
is as follows: given a new, unseen geometry, what is the likelihood that the
network prediction is within the expected accuracy based on the test set? In
other words, how can we know that a given sample is out-of-distribution (OOD)?
A solution for this might be to use convolutional autoencoders to use the same
images we have used for the CNNs, to learn an encoding of the geometries in the
dataset. Autoencoders used in this way have been successfully used to encode
geometries in a parameter--free way \cite{stall}. This encoding 
--- called a latent space representation --- could then
be used to determine whether or not a new geometry is OOD, and what the 
accuracy expected of the network can be. This ties in well with the present
planar projection representations of the present work, which work naturally
with convolutional autoencoder, and will be an avenue
for future investigations.

\section{Appendix}

\subsection{Dataset ablation}

\begin{figure}[h!]
    \centering
    \includegraphics[width=0.69\textwidth]{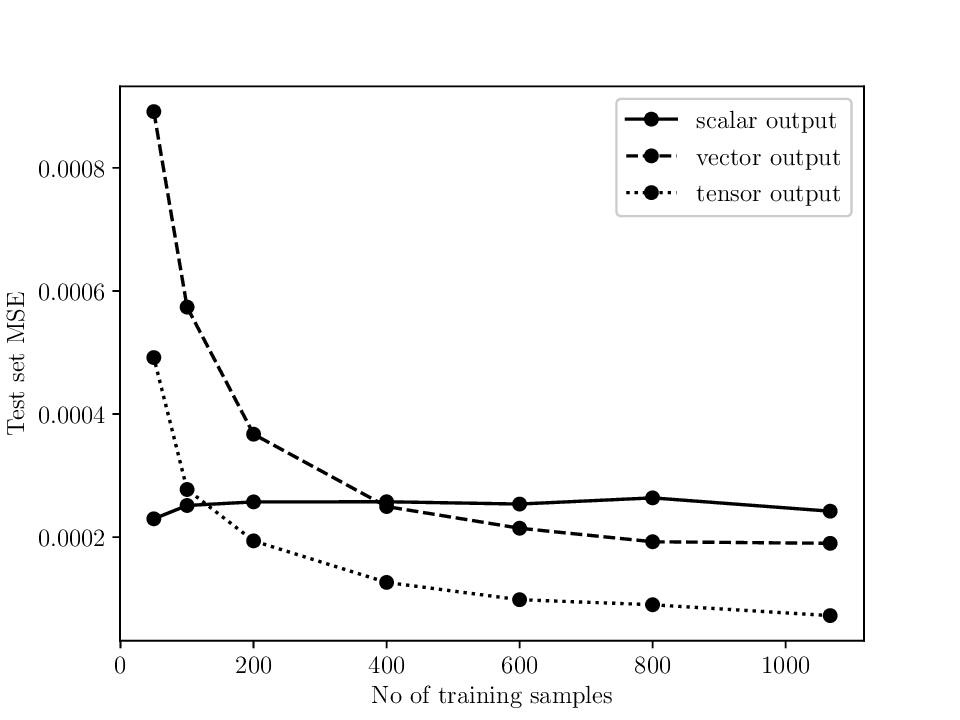}
    \caption{MSE as a function of training set size.}
    \label{f10}
\end{figure}

To get a sense of the amount of data required for this problem, we perform
ablation studies for all three models. The training set is gradually reduced
in size, keeping the validation and test sets fixed. The network architecture
is left unchanged. The results in terms of average MSE over all the test set
are shown in Figure \ref{f10}.

The models converge at all three levels of abstraction for approximately 200
samples, which corresponds to 20 samples per dimension. Below this, the tensor 
model degrades first, followed by the vector
model, as expected. 
It is important to be sparse with the sampling
in each dimension; with additional corner cases the simplex becomes higher
dimensionsal and one is quickly confronted with the curse of dimensionality
when ensuring sufficient coverage.

\subsection{Uncertainty quantification}

To establish confidence in the predictions of the model, we investigate various
sources of uncertainty, focusing on both scalar and vector cases. The method
we follow is one of ensembles of models \cite{uq}. We consider the following
two types of input uncertainties:

\begin{itemize}
    \item \textit{Aleatory uncertainty:} Associated with the randomness in the
        learning process, this is not controllable. We consider two sources:
        the randomness of the initialization of weights in the model, which can
        have a significant impact on the converged optmimum, and the selection
        of the batches used for training. The batch size is fixed at 64 after
        a hyperparameter tuning process, but there is still randomness 
        associated with the elements in each batch, which affects the gradient
        computed at each step, and consequently the optimum. 
    \item \textit{Epistemic uncertainty:} This source of uncertainty is due
        to assumptions made in the model, and here we consider the choices 
        made in the model architecture. We are interested in the impact that
        a perturbation to the size of the network can have, so we consider
        variations in the number of convolutional layers. Another important 
        source of uncertainty is the impact that adding dropout layers can 
        have, since this indicates how perturbations to the different 
        connections in the network can impact the predictions. We note that 
        the amount of dropout that is generally needed is much lower in 
        convolutional networks (as opposed to fully-connected ones), owing to
        the need for the identification of local features, and the fact that
        connections are sparser to begin with.
    \end{itemize}

Table \ref{t3} summarizes the variation in the ensemble of models. We follow
the recommendations of Saetta \textit{et al.} \cite{st} in sizing the ensemble;
they found that for a problem of similar scope, an ensemble size of 50 was 
sufficient to converge the uncertainty bounds. In the present work, we use 90
samples to build the ensemble.

\begin{table}[!htb]
\centering
\caption{Details of UQ ensembles}
\begin{tabular}{lll}
\hline
Parameter              & Uncertainty type & Range      \\ \hline
Weights initialization & Aleatory         & n/a (seed) \\
Batch selection        & Aleatory         & n/a (seed) \\
Number of Conv layers  & Epistemic        & 3 to 5     \\
Dropout \%             & Epistemic        & 0.5 to 2   \\ \hline
\label{t3}
\end{tabular}
\end{table}

\begin{figure}[h!]
    \centering
    \includegraphics[width=0.69\textwidth]{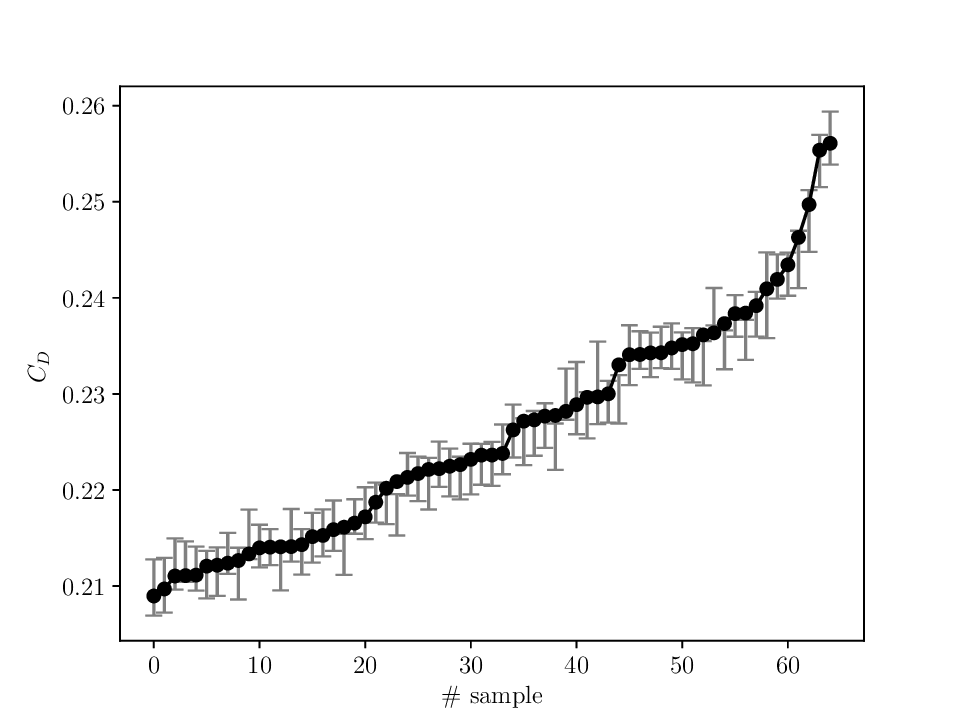}
    \caption{Scalar model 100\% CI on test set. Black dots represent
    ground truth values.}
    \label{f15}
\end{figure}

\begin{figure}[h!]
    \centering
    \includegraphics[width=0.69\textwidth]{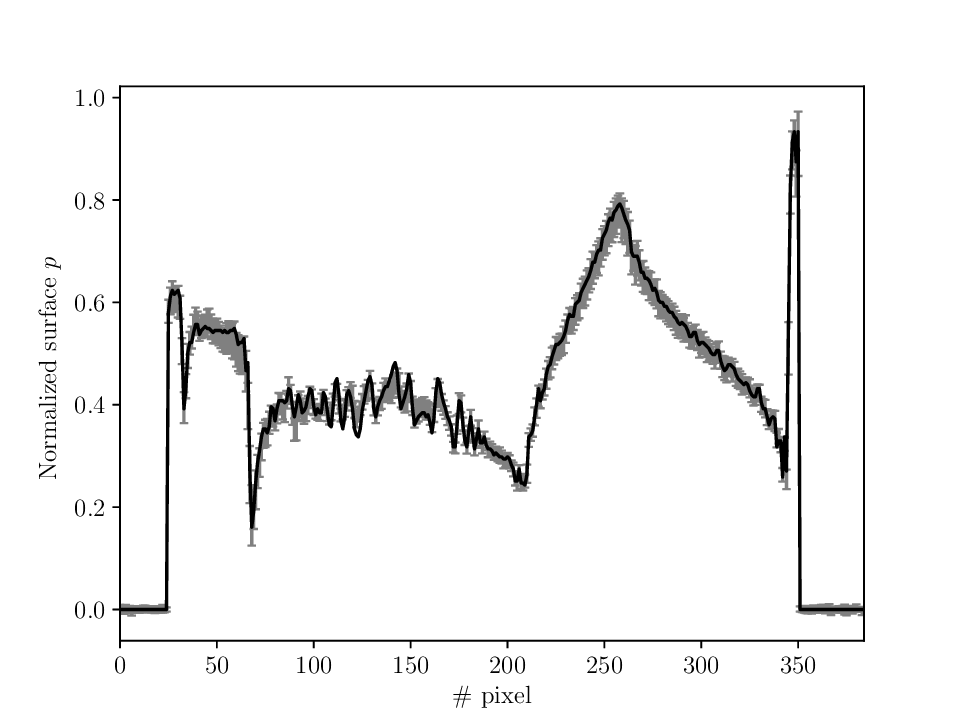}
    \caption{Scalar model 100\% CI on test set. Black line represents
    ground truth values.}
    \label{f22}
\end{figure}

Figure \ref{f15} shows the 100\% confidence intervals of the models for all
points in the testing set. We observe that there is a scatter of about 30 counts
across the set (95th percentile), which is consistent with the accuracy of 
a single model with respect to the ground truth values. We also perform this
ensemble analysis for the vector case (Figure \ref{f22}), and see that there is
an increased scatter near the back of the car between pixels 50 nad 150, as
we expect. 

\section*{Data availability statement}
All data used in this study will be made available upon request. The software 
developed in this work is released as an open-source project on Github.

\section*{Acknowledgments}

Computations performed in this work were performed on the Kestrel Computing 
System at the National Renewable Energy Lab, Golden, CO. The authors gratefully
acknowledge useful discussions with Frank Ham, Sanjeeb Bose and Christopher Ivey.

\bibliographystyle{plain}  
\bibliography{r}  

\end{document}